\documentclass[pdflatex,sn-vancouver-num]{sn-jnl}% Vancouver Numbered Reference Style
\usepackage{graphicx}%
\usepackage{multirow}%
\usepackage{amsmath,amssymb,amsfonts}%
\usepackage{amsthm}%
\usepackage{mathrsfs}%
\usepackage[title]{appendix}%
\usepackage{xcolor}%
\usepackage{textcomp}%
\usepackage{manyfoot}%
\usepackage{booktabs}%
\usepackage{algorithm}%
\usepackage{algorithmicx}%
\usepackage{algpseudocode}%
\usepackage{listings}%
\usepackage{mdframed}
\usepackage{tikz}
\usetikzlibrary{positioning, arrows.meta}

\theoremstyle{thmstyleone}%
\theoremstyle{thmstyletwo}%

\theoremstyle{thmstylethree}%

\begin{document}

\title{Differentiable latent structure discovery for interpretable forecasting in clinical time series}

\author*[1,2]{\fnm{Ivan} \sur{Lerner}}\email{ivan.lerner@u-paris.fr}

\author[1]{\fnm{Jean} \sur{Feydy}}

\author[2]{\fnm{Alexandre} \sur{Kalimouttou}}

\author[2]{\fnm{Anita} \sur{Burgun}}

\author[3,4,5]{\fnm{Francis} \sur{Bach}}

\affil*[1]{\orgname{Université Paris Cité and Université Sorbonne Paris Nord, INRIA, Inserm, HEKA}, 
\orgaddress{\city{Paris}, \postcode{75015}, \country{France}}}

\affil[2]{\orgname{Department of Medical Informatics, Assistance Publique Hôpitaux de Paris (AP-HP), Georges Pompidou European Hospital}, 
\orgaddress{\city{Paris}, \country{France}}}

\affil[3]{\orgname{SIERRA, Inria Paris}, 
\orgaddress{\city{Paris}, \postcode{75015}, \country{France}}}

\affil[4]{\orgname{Département d’Informatique de l’École Normale Supérieure}, 
\orgaddress{\country{France}}}

\affil[5]{\orgname{PSL Research University}, 
\orgaddress{\city{Paris}, \country{France}}}


\abstract{\footnote{This manuscript is under review at BioData Mining.}
\textbf{Background:}
Timely, uncertainty-aware forecasting from irregular electronic health records (EHR) can support critical-care decisions, yet most approaches either impute to a grid or sacrifice interpretability.
We introduce StructGP, a continuous-time multi-task Gaussian process that couples \emph{process convolutions} with \emph{differentiable structure learning} to uncover a sparse, ordered directed acyclic graph (DAG) of inter-variable dependencies while preserving principled uncertainty.
We further propose LP-StructGP, which augments StructGP with \emph{latent pathways}—shared, temporally shifted trajectories inferred via subject-specific coupling filters and a softmax gating mechanism—to capture cross-patient progression patterns.
Both models are trained under sparsity and acyclicity constraints (augmented Lagrangian, Adam) using scalable low-rank updates: StructGP via exact marginal likelihood maximization and LP-StructGP via an online conditional marginal-likelihood objective.

\textbf{Results:}
In controlled, correctly specified simulations, graph recovery improved with
cohort size, with the median Structural Hamming Distance reaching zero at the
largest cohort size, while pathway assignments showed high Adjusted Rand
Index. Additional experiments showed reduced graph-recovery performance under
MIMIC-IV-derived observation schedules.
Our analysis establishes that the ordered StructGP graph is identifiable from
the population marginal likelihood under the stated model, observation-design,
and noise assumptions. For LP-StructGP, the corresponding result identifies
the shared ordered inter-task graph conditionally on fixed subject-level
pathway filters.
On a MIMIC-IV septic shock cohort (n{=}1{,}008; norepinephrine, creatinine, mean blood pressure), StructGP improves short-horizon (6\,h) forecasting over independent-task baselines (average RMSE 0.68 [95\% CI: 0.63--0.74] vs. 0.88 [0.83--0.94]) and, with 15 additional inputs, markedly outperforms unstructured kernels (0.63 [0.58--0.69] vs. 3.02 [2.85--3.18]) with superior calibration (coverage 0.96 vs. 0.84).
For long horizons (up to 6\,days), LP-StructGP further reduces error for creatinine (RMSE 0.95 [0.88--1.03] vs.\ 1.17 [1.08--1.25]) and improves overall coverage (0.93 [0.93--0.94] vs.\ 0.91 [0.91--0.92]).
On the PhysioNet Challenge (12k patients, 41 variables), StructGP attains competitive accuracy (MAE $3.72{\times}10^{-2}$) relative to a strong published comparator graph neural model.

\textbf{Conclusion:}
These results show that structured process convolutions with latent pathways deliver interpretable, scalable, and well-calibrated forecasting for irregular clinical time series.
}




\maketitle
\section{Introduction}\label{sec:intro}

Clinical decision-making relies not only on assessing a patient’s current state but also on anticipating its future course.
Care plans, for example, are adjusted when renal function declines, and more aggressive interventions may be required if cardiovascular stability cannot be restored.
Predictive models that provide model-based, data-driven forecasts can therefore play a crucial role in supporting clinicians.
Their potential impact is considerable: up to 11\% of deaths in critically ill patients are attributable to delayed care~\cite{thomson2007safer}, underscoring the importance of timely and accurate predictions.

Electronic health records (EHRs) capture a wide range of information during hospital admissions, including sensor measurements, biomarkers, and treatments.
Excluding unstructured text and imaging data, these records can be viewed as collections of multivariate time series, one per patient's stay.
Unlike controlled experimental settings, clinical observations are collected opportunistically rather than on a regular grid, making them difficult to model with classical frameworks (e.g., vector autoregressive models).
The resulting patterns are irregular and heterogeneous (Supplementary Figure S2), with variables measured at different times, frequencies, and subsets of patients.

A common strategy to address irregular sampling is to resample the data onto a fixed grid and impute the missing values, after which standard time-series methods can be applied~\cite{lu2018hierarchicalvar, PURUSHOTHAM2018112}.
While simple, this approach is computationally intensive and introduces potential biases through resampling and imputation.
A more principled alternative is to work directly in continuous time, thereby avoiding these intermediate steps.
Continuous-time modeling can be approached through a variety of frameworks: multi-task Gaussian processes (MTGPs)~\cite{durichen2014multitask, ghassemi2015multivariate, ZHANG2022104079, liu2013modeling, liu2016learning, cheng2020sparse, cui2021hierarchical, karch2020gaussian}; neural ODEs and related stochastic differential equation models~\cite{rubanova2019latent, de2019gru, scholz2023latent, bilovs2021neural}; adaptations of deep learning architectures, such as recurrent~\cite{che2018recurrent} or attention-based networks~\cite{zhang2023crossformer, zhang2024irregular, liu2023itransformer, zhang2023warpformer}; and, more recently, graph neural networks (GNNs)~\cite{wu2020connecting, cao2020spectral, yalavarthi2024grafiti}.
Although GNN-based methods offer some interpretability and, in the case of GraFITi~\cite{yalavarthi2024grafiti}, state-of-the-art predictive performance, they generally lack principled uncertainty quantification, a key strength of MTGPs.

In this work, we revisit the classical framework of process convolutions~\cite{boyle2005dependent, higdon2002space, alvarez2013linear}, which defines the covariance structure of a multi-task Gaussian process (MTGP) through kernel convolutions.
We combine this framework with modern differentiable structure learning~\cite{zheng2018dags, pamfil2020dynotears}, enabling the discovery of interpretable dependency structures between time series in continuous time.
A preliminary version of this approach was presented at the Bayesian Decision-making and Uncertainty Workshop (NeurIPS 2024)~\cite{lerner2025spectral}.

The present work extends it by introducing internal standardization, latent pathways, an online learning algorithm, an extensive biomedical evaluation,  and a population identifiability analysis of the ordered inter-task graph.
In our formulation, the sparsity pattern of the cross-task convolution filters is represented as a directed acyclic graph (DAG), encoding ordered conditional independence relations among time series.
The corresponding adjacency matrix and variable ordering can be inferred directly from data by solving a continuous optimization problem~\cite{zheng2018dags, pamfil2020dynotears}.
Compared to classical MTGPs, our approach provides both flexible probabilistic modeling and interpretable dependency structures between time series.

In addition, we address the question of modeling dependencies across patients by introducing the concept of a latent pathway: a partially shared trajectory that groups of patients may follow.
This assumption is broad enough to capture diverse phenomena, ranging from different patients experiencing the same underlying pathological process (e.g., urinary tract– versus lung infection–induced sepsis) to patients undergoing similar treatment strategies (e.g., progressive catecholamine withdrawal).
At test time, our model infers both the latent pathway along which a patient is evolving and their position within that pathway.
In contrast to previous approaches that tackled patient-level dependencies~\cite{liu2016learning, cheng2020sparse, cui2021hierarchical}, we integrate latent pathways directly into the covariance design.
This yields calibrated uncertainty estimates while preserving interpretable dependency structures across tasks.

With this study, our objectives were to design, implement, and evaluate a process convolution model, StructGP, along with an extension that incorporates latent pathways, LP-StructGP, together with their associated learning algorithms.
We assessed their properties using both controlled simulations, including experiments using empirical
MIMIC-IV-derived observation schedules, and real-world data, specifically a MIMIC-IV septic shock cohort from a previous study~\cite{kalimouttou2023machine, johnson2020mimic}.

For comparison with state-of-the-art deep learning methods, we further evaluated our models on the PhysioNet Challenge~\cite{silva2012predicting}, using the preprocessing pipeline described in~\cite{zhang2024irregular}.
Our contributions are thus twofold: (i) StructGP, which unifies process convolutions with differentiable structure learning, and (ii) LP-StructGP, which extends this framework to capture patient-level latent pathways.

\section{Methods}\label{sec:methods}

\subsection{Process convolution models}\label{subsec:process_convolution}

We begin by recalling the process convolution formulation of Gaussian processes~\cite{rasmussen2003gaussian}, which provides a flexible framework for constructing structured covariance functions.  
Let $y(\mathbf{x}) = \{y(\mathbf{x}) : \mathbf{x} \in \mathcal{X}\}$ denote a stochastic process, i.e., a collection of random variables indexed by $\mathbf{x} \in \mathcal{X}$.
Process convolution models~\cite{boyle2005dependent, higdon2002space} adopt a signal-processing perspective, where $y(\mathbf{x})$ is viewed as the output of a linear time-invariant (LTI) system driven by a latent noise process.
An LTI system is entirely characterized by its impulse response function $h(\mathbf{x})$, which represents the system’s response to a Dirac input.
If the input is a Gaussian white noise process $w(\xi)$ with covariance $\mathbb{E}[w(\xi)w(\xi')] = \delta(\xi - \xi')$, then by the closure of Gaussian processes under linear operations, $y(\mathbf{x})$ is also a Gaussian process provided that $h$ has compact support or is square-integrable:
\begin{align}
    y(\mathbf{x}) &= \int_{-\infty}^{\infty} h(\mathbf{x} - \xi) w(\xi)\, d\xi = (h * w)(\mathbf{x}), \\
    k(\mathbf{x}, \mathbf{x}') &= \int_{-\infty}^{\infty} h(\mathbf{x} - \xi) h(\mathbf{x}' - \xi)\, d\xi = (h * h)(\mathbf{x} - \mathbf{x}'),
\end{align}
where $k(\mathbf{x}, \mathbf{x}')$ denotes the covariance function of $y(\mathbf{x})$.
By appropriately designing these filters, one can derive analytical covariance functions that naturally capture the smoothness and cross-dependence of observed clinical time series, while remaining well-defined under irregular sampling.

\subsection{Structured process convolution models}\label{subsec:structured_convolution}

\subsubsection{StructGP: an inter-task graphical structure}\label{subsubsec:inter_task_structure}

We now consider \textit{StructGP}, a $k$-dimensional Gaussian process $\mathbf{Y}(t)$ obtained by convolving a multivariate white noise process $\mathbf{w}(t)$ with a matrix-valued filter $\mathbf{H}(t)$:
\begin{align}
    \mathbf{Y}(t) &= (\mathbf{H} * \mathbf{w})(t), \\
    \mathrm{Cov}[\mathbf{Y}(t), \mathbf{Y}(t')] &= (\mathbf{H} * \mathbf{H}^{\top})(t - t'),
\end{align}
where $\mathbf{w}(t)$ is a $k$-dimensional Gaussian white noise process.
The filter $\mathbf{H}(\Delta t)$ is a sparse, lower-triangular,
matrix-valued impulse response function, where $\Delta t$ denotes a time lag.
This formulation generalizes the univariate process convolution to a structured multi-task setting.

The notion of conditional independence between stationary Gaussian process components can be characterized in the frequency domain through their spectral density matrix~\cite{eichler2012fitting, brillinger1996remarks, dahlhaus2000graphical, bach2004learning}.  
In our previous work~\cite{lerner2025spectral}, we showed how this concept enables learning ordered conditional relations between time series directly from data.  
Accordingly, we parameterize the filter matrix as
\begin{align}
    \mathbf{H}(\Delta t)
    = (\mathbf{I}+\mathbf{S})\circ\mathbf{L}(\Delta t),
\end{align}
where $\circ$ denotes the Hadamard product, $\mathbf{S}$ is a sparse lower-triangular matrix (up to permutation) encoding inter-task dependencies, and $\mathbf{L}(\Delta t)$ is a positive matrix of filter coefficients with elements
\begin{align}
    L_{vu}(\Delta t)
    = \exp\!\left(-\frac{(\Delta t)^2}{\ell_{vu}}\right),
\end{align}
with $\ell_{vu} > 0$ denoting learnable length scales.  

As shown in our previous work~\cite{lerner2025spectral}, the sparsity pattern
of the triangular spectral factor encodes ordered conditional-independence
relations between stationary Gaussian process components. After permutation
into the learned task ordering, for each pair $u<v$,
\begin{align}
    \mathbf{S}_{vu} \neq 0
    \quad \Leftrightarrow \quad
    Y_u \rightarrow Y_v
    \quad \Leftrightarrow \quad
    Y_u \not\!\perp\!\!\!\perp Y_v
    \mid \mathbf{Y}_{1:u-1}.
\end{align}
Here, $\mathbf{Y}_{1:u-1}$ denotes the collection of process trajectories
preceding $Y_u$ in the learned ordering. Thus, $\mathbf{S}_{vu}=0$ means that
$Y_u$ and $Y_v$ are conditionally independent after accounting for these
preceding processes. The resulting distribution satisfies the Markov
factorization associated with the ordered graph~\cite{lerner2025spectral}.
Each filter element
$H_{vu}(\Delta t)$ parameterizes the temporal profile of the corresponding
model-based dependency. The arrow represents the learned ordering and does
not, by itself, imply a causal direction.

Additional file~1, Section~S3.1 establishes population identifiability of this
ordered-dependence graph within the anchored acyclic Gaussian-filter StructGP
class. The proof works directly with the process-convolution spectral factor:
the fixed unit diagonal self-filter amplitudes and acyclicity make this factor,
and hence its off-diagonal support, unique. Combined with a sufficiently rich
observation design and an observation-noise term that is absent, known, or
separately identifiable, this yields graph identifiability from the population
marginal likelihood. This is a population statement and does not by itself
guarantee finite-sample graph recovery or successful optimization of the
nonconvex objective.

To reduce sensitivity to marginal scale artifacts, we introduce a signal-to-noise-ratio (SNR)-preserving internal standardization step inspired by the internally standardized structural causal model (iSCM) of Ormaniec \textit{et al.}~\cite{ormaniec2024standardizing}, but adapted to the process-convolution setting.
The standardization is defined in Additional file 1, Section S1.2.
Whereas iSCMs standardize the observed variables, StructGP standardizes the latent process-convolution component of each task to unit marginal variance.

Under this row-wise standardization, $\widetilde S_{vu}$ represents the peak
filter amplitude relative to the total latent variability of target task $v$,
rather than an absolute amplitude in its original measurement units. A
multiplicative rescaling of task $v$ rescales both $S_{vu}$ and its
normalization factor by the same amount, leaving $\widetilde S_{vu}$
unchanged. This is desirable because absolute amplitudes depend on arbitrary
measurement units. When filter lengthscales differ, the relative contribution
to marginal variance depends on both amplitude and lengthscale, as detailed in
Additional File~1, Section~S1.2.

In the SNR-preserving version, the same taskwise scale is applied to a shared
raw observation-noise variance, so that task-specific signal-to-noise ratios
are preserved. Additional file~1, Section~S3.2 shows that the standardized
observation law retains population identifiability of the ordered StructGP
graph when the raw noise variance is shared and positive and the observation
design separates the continuous latent covariance from the nugget term.

\subsubsection{LP-StructGP: an inter-subject latent pathway structure}\label{subsubsec:inter_subject_structure}

\begin{figure}[hbt!]%% placement specifier
\centering%% For centre alignment of image.
\includegraphics[width=\textwidth]{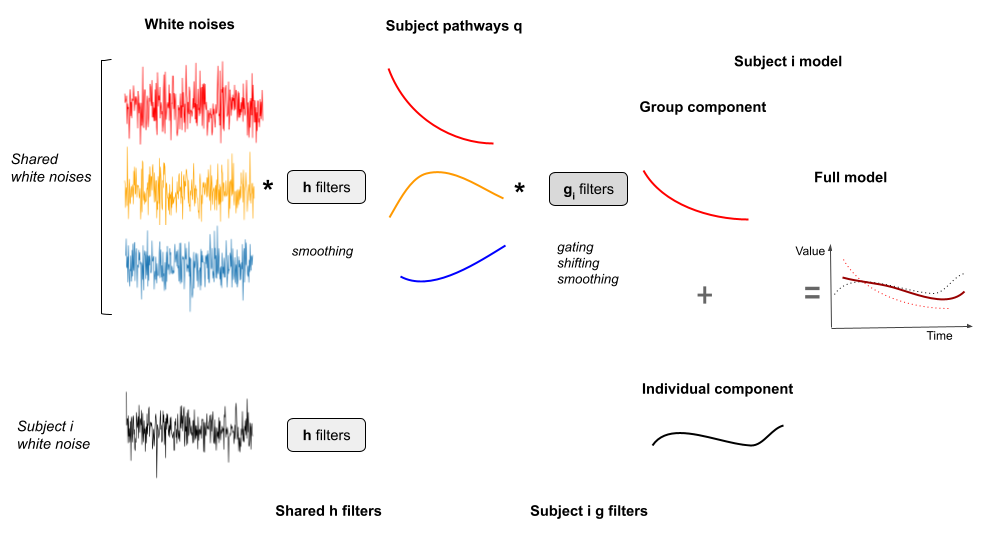}
\caption{Sketch of the latent pathway model for one task\\
  }\label{fig:lp_kernel}
\end{figure}

To capture shared temporal dynamics across subjects, we extend StructGP with an additional latent layer representing group-level trajectories, referred to as \textit{latent pathways} (see Figure \ref{fig:lp_kernel}).  
Each latent pathway corresponds to a multivariate Gaussian process
\begin{align}
    \mathbf{Q}_u(t) &= (\mathbf{H} * \mathbf{w}_u)(t),
\end{align}
where $\mathbf{w}_u(t) \in \mathbb{R}^k$ is a latent multivariate white noise process and $\mathbf{H}(t)$ is the shared inter-task filter.  
Individual trajectories are modeled as mixtures of these latent pathways, together with a subject-specific component:
\begin{align}
    \mathbf{Y}_i(t) 
    &= \sum_{u=1}^p (\mathbf{G}_{iu} * \mathbf{Q}_u)(t) + (\mathbf{H} * \mathbf{w}_i)(t),
\end{align}
where $\mathbf{G}_{iu}(t)$ is a subject-specific coupling filter.  
To account for differences in disease progression, $\mathbf{G}_{iu}(t)$ includes a temporal shift parameter, while subject membership to each pathway is controlled through a softmax-based gating mechanism~\cite{jacobs1991adaptive, shazeer2017outrageously}.  
This formulation allows subjects to be probabilistically assigned to shared latent trajectories while preserving subject-specific variability.  
Full mathematical details and the corresponding covariance expressions are provided in Additional file 1, Section S1.3.

Because the subject-level pathway filters are scalar and act identically
across tasks, they multiply the shared task spectral density by a strictly
positive frequency-dependent scalar and therefore preserve the support of its
ordered spectral factor. As shown in Additional file~1, Section~S3.3, the
shared ordered inter-task graph is identifiable from the population marginal
likelihood when the same subject-level pathway filters are held fixed across
competing parameterizations, the observation design is sufficiently rich, and
the observation-noise term is absent, known, or separately identifiable. This
is a conditional graph-identifiability result: it does not establish joint
identifiability of the shared filter together with the pathway gates,
lengthscales, or temporal shifts.

\subsection{Learning and inference}\label{subsec:learning_inference}
As the time series are irregularly sampled across patients, we adopt a set-based indexing scheme that treats all observations as elements of a common collection rather than as regularly spaced sequences.
Owing to the marginalization properties of multivariate Gaussians, the covariance structure defined above remains valid under this irregular sampling.
This allows us to directly apply standard Gaussian process learning and inference procedures: (i) maximizing the process marginal likelihood to learn the convolution filter parameters, and (ii) computing posterior predictive distributions at test points (see Additional file 1, Section S1.4).

\subsubsection{Differentiable structure learning}\label{subsubsec:diff_structure_learning}

To uncover interpretable dependencies between tasks, we learn the adjacency matrix $\mathbf{S}$ under the constraint that it forms a weighted directed acyclic graph (DAG) $\mathcal{G}$.
We adopt the differentiable structure learning framework of NOTEARS~\cite{zheng2018dags}, which enforces acyclicity through a smooth matrix function based on the trace of the matrix exponential (see Additional file 1, Section S1.5.1).

The parameters $\theta = \{\mathbf{S}, \mathbf{L}\}$ are estimated by minimizing the process negative marginal likelihood with sparsity and acyclicity constraints:
\begin{align}
  \theta^* = \operatorname*{argmin}_\theta
  \Big[-\log \mathcal{L}(\mathbf{y}, \mathbf{X}, \theta)
  + \mathcal{P}_{\lambda}(\mathbf{S})\Big]
  \quad
  \text{s.t. } h(\mathbf{S}) = 0,
\end{align}
where $\mathcal{P}_{\lambda}$ is a sparsity penalty and $h(\mathbf{S})$ enforces the DAG constraint.
Optimization is carried out with the augmented Lagrangian method~\cite{nemirovsky1999optimization}, using Adam~\cite{kingma2014adam} as the gradient-based optimizer.
A smooth approximation of the $\ell_1$ norm ensures differentiability everywhere, and the sparsity weight $\lambda$ is selected by minimizing an AIC-like or validation criterion.
Finally, a hard threshold is applied to $\mathbf{S}$ to obtain an acyclic structure.
Full algorithmic details are provided in Additional file 1, Section S1.5.

Acyclicity serves structural and identifiability roles distinct from
forecasting. It is not required for $\mathbf{H}*\mathbf{H}^{\top}$ to define a
valid covariance, but it makes the support of $\mathbf{H}$ lower triangular up
to permutation, yielding an ordered DAG with a conditional-dependence
interpretation. It also enables the recursive root-based recovery used in our
population identifiability proof. Together with sparsity, acyclicity reduces
the effective parameter space and may therefore act as predictive
regularization.

\subsubsection{Low-rank plus block-sparse scheme for scalability}\label{subsubsec:lowrank_blocksparse}

A key limitation of Gaussian process regression is its $O(n^3)$ computational complexity, with $n$ the number of observations, due to the covariance matrix inversion.
To scale to large cohorts, we exploit independence across subjects and implement all models with batched Cholesky solvers in PyTorch, leveraging GPU-based parallel computation.
The resulting complexity of matrix inversion scales with the number of observations per subject, making this approach efficient in regimes with many subjects but fewer than $\sim$1000 observations per subject.
Independence across subjects further allows scaling through mini-batch gradient steps (e.g., using the Adam optimizer).

In \textsc{StructGP}, subjects are independent by design, and mini-batches are
constructed by padding subjects with different sequence lengths.
By contrast, \textsc{LP-StructGP} couples subjects through shared latent
pathways. To make this tractable, we approximate the inter-subject covariance
with the Hilbert-space Gaussian process (HSGP) method
\cite{solin2020hilbert,riutort2023practical} (see Additional file 1,
Section S1.6). This yields a low-rank plus block-sparse covariance
\[
    K_\theta = M_\theta+\Phi_\theta\Phi_\theta^\top,
\]
where \(M_\theta\) is block-diagonal over subject-level blocks and
\(\Phi_\theta\) contains the low-rank features associated with the shared
latent pathways. Inference can then be performed using the Woodbury identity
and the matrix determinant lemma, requiring only subject-level Cholesky
factorizations and a low-rank solve of dimension \(mkp\), where $m$ is the number of basis functions in the HSGP approximation, $k$ is the number of tasks, for \(p\) latent pathways.

For mini-batch training, we use an online conditional marginal likelihood
based on cumulative Woodbury increments (Additional file 1, Sections
S1.7--S1.8). The accumulated low-rank statistics define a Gaussian
state over the finite-dimensional shared pathway coefficients. In the
no-fading case, with fixed parameters during a full sweep, the batchwise
quadratic and log-determinant increments telescope and recover the exact
finite-rank HSGP NMLL. In practice, previously accumulated statistics are
detached from the current automatic-differentiation graph and may be
geometrically discounted, yielding a scalable detached, fading-memory approximation to this reference objective.
Standard HSGP results control the approximation of the finite-rank posterior
to its exact GP counterpart, whereas statistical recovery of the true latent
pathways additionally requires identifiability and increasing information in
the relevant latent directions. These conditions, together with the stability
of the detached accumulator, are discussed in Additional file 1,
Section S3.4.

\subsection{Simulation study}\label{subsec:simulation}

We evaluated the ability of StructGP and LP-StructGP to recover both the inter-task graph and the latent pathway structure from simulated data.
For each simulation, a ground-truth graph, covariance parameters, and observations were sampled from the corresponding generative model (see \ref{subsec:structured_convolution}).
Models were then fitted using the learning procedure described in \ref{subsec:learning_inference}, and the inferred structures were compared to the ground truth.

In Experiment A, inter-task graph recovery was evaluated using the Structural Hamming Distance (SHD), defined as the number of edge insertions, deletions, or reversals required to recover the true graph, and the edge-level F1 score. Latent pathway recovery was evaluated using the Adjusted Rand Index (ARI).
Experiment A used 10 tasks, 25 observations per task, Erd\H{o}s--R\'enyi
graphs with mean degree~2, and an observation-noise variance of
$\sigma^2=10^{-2}$ provided during fitting; for LP-StructGP, the number of
latent pathways was varied over $\{3,6,12\}$.
Each configuration was repeated 30 times, and results are summarized by the median and interquartile range.
Experiment B used the same setting as Experiment A for StructGP with 1,000 subjects, but incorporated realistic observation patterns by sampling, with replacement, entire timestamp sequences from observed MIMIC-IV time series.
This configuration was repeated 150 times, and F1 scores were stratified by quartiles of the mean number of observations per observed pair.
In Experiment C, we evaluated whether the true number of latent pathways could be recovered by selecting the model with the lowest NMLL in a high shared-to-individual component ratio setting. Oracle inter-task kernel parameters were fixed and 100 subjects simulated to reduce computational cost. Each true pathway count was evaluated over 50 repetitions, and we report the mean selected number of pathways with 95\% bootstrap confidence intervals.

Full simulation settings, parameter sampling schemes, and additional results are provided in Additional file 1, Section S1.9 and in Lerner et~al.~(2024)~\cite{lerner2025spectral}.

\subsection{Real-life evaluation}\label{subsec:real_life_eval}

\subsubsection{MIMIC-IV septic shock cohort}\label{subsubsec:mimic}

We evaluated StructGP and LP-StructGP on a septic shock cohort extracted from the MIMIC-IV database~\cite{johnson2020mimic, kalimouttou2023machine}.
MIMIC-IV is a publicly available collection of de-identified ICU records from the Beth Israel Deaconess Medical Center (Boston, US) between 2008 and 2019.
Septic shock was defined according to Sepsis-3 criteria~\cite{singer2016third}.
Only the first ICU stay per patient was retained.
Detailed inclusion criteria and preprocessing steps are provided in Additional file 1, Section S1.10.

We aimed to jointly forecast creatinine (Cr), norepinephrine (NE) dose, and mean blood pressure (MBP) over the first seven ICU days.
Two forecasting regimes were considered: (\textit{i}) short-term, where within each 24-hour window the model predicts the last 6 hours from the preceding 18 hours of observations, and (\textit{ii}) long-term, predicting the remainder of the stay from the first 24~hours.
All variables were standardized using a quantile transform fitted on the training set.
Data were split by subject into training, validation, and test sets (70\%/15\%/15\%).

We compared kernel architectures (StructGP, LP-StructGP, and an independent-task baseline) under identical hyperparameter settings.
The \emph{Independent Tasks} baseline sets all off-diagonal entries of the
inter-task filter matrix to zero, resulting in independent task-specific GPs
with no cross-task covariance. The \emph{No Structure} baseline uses the same
multivariate process-convolution kernel as StructGP, but its inter-task filter
matrix is learned without the sparsity penalty or acyclicity constraint.
Consequently, the comparison between No Structure and StructGP isolates the
effect of these structural constraints.

Performance was assessed using the Root Mean Square Error (RMSE) and the empirical 95\% coverage of the predictive distribution.
Additional preprocessing and feature-engineering details are given in Additional file 1, Section S1.10.

To assess scalability and multi-task generalization, we conducted additional large-scale experiments including 18 physiological and treatment variables spanning cardiovascular, respiratory, renal, and metabolic systems (see Additional file 1, Section S1.10 for the complete list).

We compare our predictions with those of the GraFITi model~\cite{yalavarthi2024grafiti}, using the same configuration as for the PhysioNet dataset, namely 4 layers, 2 attention heads, and a latent dimension of 64.

\subsubsection{PhysioNet Challenge dataset}\label{subsubsec:physionet}

We further evaluated StructGP on the PhysioNet Challenge dataset~\cite{silva2012predicting}, following the preprocessing pipeline of Zhang~et~al.~\cite{zhang2024irregular}.
The dataset includes approximately 12{,}000 ICU patients, each with 41 irregularly sampled clinical variables recorded during the first 48~hours after admission.
For each patient, the first 24~hours were used as observations to predict values over the following 24~hours.
Data were split into training, validation, and test sets (60\%/20\%/20\%).
For comparison, we also refitted the GraFITi model~\cite{yalavarthi2024grafiti}, which currently achieves a strong published performance on this benchmark.

\clearpage
\section{Results}\label{sec:results}

\subsection{Simulation study}\label{res:simulation}

\begin{figure}[hbt!]
  \centering
  \includegraphics[width=1.\textwidth]{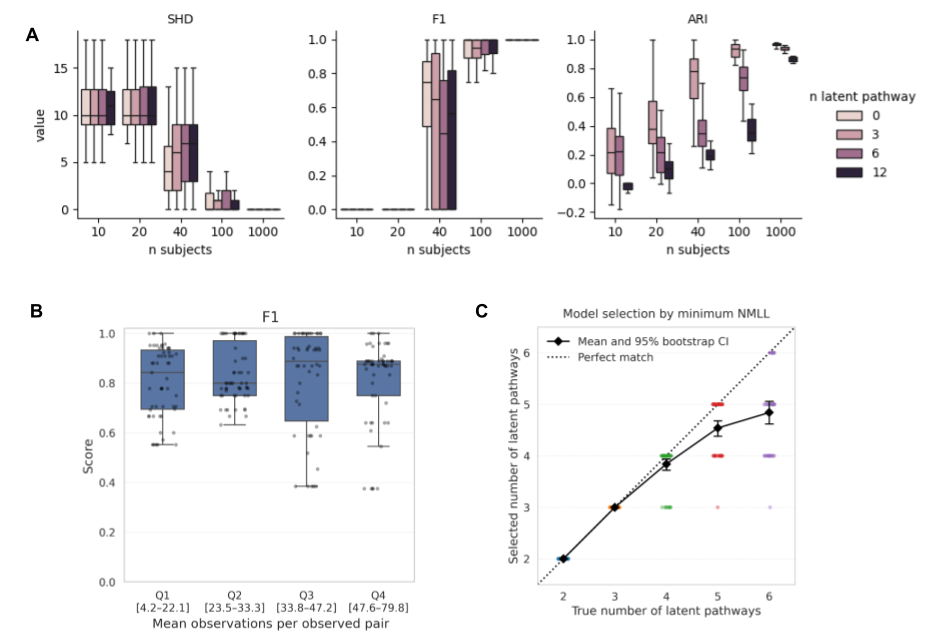}
\caption[Recovery of graph and latent pathway structure in simulation]{Recovery of graph and latent pathway structure in simulation.\\
\footnotesize
\textbf{A}, Inter-task graph recovery, measured by Structural Hamming Distance (SHD) and edge-level F1 score, and latent pathway recovery, measured by Adjusted Rand Index (ARI), across sample sizes and numbers of latent pathways. Boxplots summarize 30 repetitions per configuration.
\textbf{B}, Edge-level F1 scores with 1,000 subjects, no latent pathways, and realistic observation patterns generated by sampling, with replacement, entire timestamp sequences from MIMIC-IV time series. Results from 150 repetitions are stratified by quartiles of the mean number of observations per observed pair.
\textbf{C}, Number of latent pathways selected by minimum NMLL versus the true number under a high shared-to-individual component ratio regime. Points and error bars show the mean and 95\% bootstrap confidence interval across 50 repetitions; the dotted line indicates perfect recovery.
Simulation parameters are detailed in Additional File~1, Section~S1.9.
F1 denotes the harmonic mean of precision and recall for edge recovery.}
  \label{fig:sim_tsm}
\end{figure}

In Experiment A (Figure~\ref{fig:sim_tsm}, Panel A), under correct model
specification, uniform random sampling, low observation noise, and an oracle
noise variance, recovery of the inter-task graph and latent-pathway assignments
improved with cohort size. At 1,000 subjects, the median SHD
reached zero and agreement between inferred and true pathway assignments was
high.
In Experiment B (Figure~\ref{fig:sim_tsm}, Panel B), replacing the uniform
design with empirical MIMIC-IV-derived observation schedules reduced
edge-recovery performance, although F1 scores remained high in the setting
considered.
In Experiment C (Figure~\ref{fig:sim_tsm}, Panel C), under a high
shared-to-individual component ratio and with oracle inter-task kernel
parameters, NMLL-based model selection recovered the number of latent pathways
when sufficient subjects represented each pathway. Selection became downward
biased when the mean number of subjects per pathway fell below approximately
25, corresponding in this experiment to more than four latent pathways.

Supplementary Figure S3 shows how the predictive mean initially combines subject-specific and group-level components, before converging to the group-level forecast at longer horizons in a simulated cohort.

\subsection{MIMIC-IV septic shock cohort}\label{res:mimic}
We evaluated short-term forecasting on a MIMIC-IV septic shock cohort comprising 1,008 ICU stays, split into training (705), validation (151), and test (152) subjects.
Each record included three target variables—norepinephrine administration rate (34,411 observations), serum creatinine (8,959), and mean blood pressure (109,635)—measured over the first seven ICU days (time window 0–7~days).
Measurements were sampled with a minimum sampling interval of one hour.
When including 15 additional physiological and treatment variables, the dataset totaled 511,186 observations.
All time series were standardized after preprocessing as described in Section~\ref{subsubsec:mimic}.

\subsubsection{Short-term forecast}

\begin{table}[hbt!]
    \centering
\resizebox{\textwidth}{!}{
\begin{tabular}{llllll}
\toprule
 & \textbf{Models} & \multicolumn{4}{c}{\textbf{Output Tasks}} \\
 &   & \textbf{Average} & \textbf{NE} & \textbf{Cr} & \textbf{MBP} \\
 \midrule
 &  & \multicolumn{4}{c}{ \textbf{RMSE [95\% CI]}} \\
 \multicolumn{2}{l}{\textbf{Input Tasks: NE, Cr, MBP}} \\
 & No Structure & 0.65 [0.60, 0.71] & 0.67 [0.61, 0.73] & 0.43 [0.26, 0.58] & 0.87 [0.85, 0.90] \\
 & Independent Tasks & 0.88 [0.83, 0.94] & 0.76 [0.70, 0.82] & 0.94 [0.79, 1.11] & 0.95 [0.93, 0.98] \\
 & StructGP & 0.68 [0.63, 0.74] & 0.65 [0.60, 0.71] & 0.44 [0.30, 0.58] & 0.96 [0.93, 0.98] \\
 & StructGP $+$ 2h lag features & 0.65 [0.60, 0.71] & 0.66 [0.61, 0.72] & 0.46 [0.32, 0.63] & 0.84 [0.82, 0.87] \\
 & StructGP $+$ task-spe noise & 0.63 [0.57, 0.68] & 0.58 [0.53, 0.64] & 0.43 [0.26, 0.60] & 0.86 [0.84, 0.88] \\
 & LP-StructGP & 0.64 [0.59, 0.70] & 0.64 [0.58, 0.70] & 0.40 [0.27, 0.56] & 0.88 [0.85, 0.90] \\
 \multicolumn{2}{l}{\textbf{Input Tasks: $+$ 15 tasks}} \\
 & No Structure & 3.02 [2.85, 3.18] & 2.84 [2.67, 3.03] & 3.25 [2.80, 3.77] & 2.97 [2.85, 3.09] \\
 & StructGP & 0.63 [0.58, 0.69] & 0.64 [0.58, 0.70] & 0.42 [0.28, 0.59] & 0.84 [0.81, 0.86] \\
 & StructGP $+$ 2h lag features & 0.66 [0.60, 0.72] & 0.65 [0.60, 0.71] & 0.42 [0.27, 0.59] & 0.91 [0.88, 0.93] \\
 & StructGP $+$ task-spe noise & 0.62 [0.57, 0.67] & 0.61 [0.55, 0.67] & 0.38 [0.26, 0.54] & 0.85 [0.83, 0.87] \\
 & GraFITi & 0.62 [0.56, 0.67] & 0.56 [0.51, 0.62] & 0.45 [0.28, 0.62] & 0.84 [0.82, 0.86] \\
\midrule
 &  & \multicolumn{4}{c}{ \textbf{Coverage [95\% CI]}} \\
 \multicolumn{2}{l}{\textbf{Input Tasks: NE, Cr, MBP}} \\
 & No Structure & 0.96 [0.95, 0.97] & 0.95 [0.94, 0.96] & 0.97 [0.95, 0.99] & 0.96 [0.95, 0.96] \\
 & Independent Tasks & 0.95 [0.93, 0.96] & 0.97 [0.96, 0.98] & 0.95 [0.92, 0.98] & 0.92 [0.91, 0.93] \\
 & StructGP & 0.97 [0.96, 0.98] & 0.96 [0.95, 0.97] & 0.97 [0.94, 0.99] & 0.98 [0.98, 0.99] \\
 & StructGP + 2h lag features & 0.96 [0.95, 0.97] & 0.96 [0.94, 0.97] & 0.96 [0.93, 0.98] & 0.96 [0.95, 0.97] \\
 & StructGP $+$ task-spe noise & 0.96 [0.95, 0.97] & 0.95 [0.93, 0.96] & 0.97 [0.95, 0.99] & 0.96 [0.95, 0.96] \\
 & LP-StructGP & 0.97 [0.96, 0.98] & 0.95 [0.94, 0.97] & 0.98 [0.96, 1.00] & 0.97 [0.97, 0.98] \\
 \multicolumn{2}{l}{\textbf{Input Tasks: $+$ 15 tasks}} \\
 & No Structure & 0.84 [0.82, 0.86] & 0.86 [0.84, 0.88] & 0.83 [0.77, 0.88] & 0.84 [0.82, 0.85] \\
 & StructGP & 0.96 [0.95, 0.97] & 0.96 [0.95, 0.97] & 0.97 [0.95, 0.99] & 0.96 [0.95, 0.96] \\
 & StructGP $+$ 2h lag features & 0.95 [0.94, 0.97] & 0.95 [0.94, 0.96] & 0.96 [0.92, 0.98] & 0.96 [0.95, 0.97] \\
 & StructGP $+$ task-spe noise & 0.96 [0.95, 0.97] & 0.93 [0.92, 0.95] & 0.98 [0.97, 1.00] & 0.96 [0.95, 0.97] \\
% \midrule
\bottomrule
\end{tabular}
}
% \captionsetup{width=0.99\linewidth}
\caption{Short-term (6 hour) horizon forecasting
\\\footnotesize
Root mean squared error (RMSE) and predictive coverage with bootstrap 95\% confidence intervals. 
Metrics are reported over a six-hour forecasting horizon for Norepinephrine (NE), Creatinine (Cr), and mean blood pressure (MBP), using 18 hours of observations within overlapping 24-hour ICU stay windows, with and without 15 additional tasks.
}
\label{tab:shorterm}
\end{table}

In the short-term forecasting task, StructGP outperformed the independent-task baseline, achieving an average RMSE of 0.68 [95\% CI: 0.63–0.74] compared to 0.88 [0.83–0.94] (Table~\ref{tab:shorterm}).
With three input tasks (NE, Cr, MBP), the explicit graphical structure provided no benefit over an unconstrained model (average RMSE 0.68 [0.63–0.74] vs. 0.65 [0.60–0.71]).
However, when 15 additional input tasks were included, the structured formulation became essential, yielding a markedly lower average RMSE (0.63 [0.58–0.69]) compared to the unstructured variant (3.02 [2.85–3.18]), and substantially better calibration (0.96 [0.95–0.97] vs. 0.84 [0.82–0.86]).
StructGP achieved performance comparable to GraFITi, with an average RMSE of 0.63 [0.58–0.69] versus 0.62 [0.56–0.67].
Finally, relaxing the assumption of shared observation noise and learning task-specific noise variances consistently improved performance across all settings.

\begin{figure}[hbt!]%% placement specifier
\centering%% For centre alignment of image.
\includegraphics[width=\textwidth]{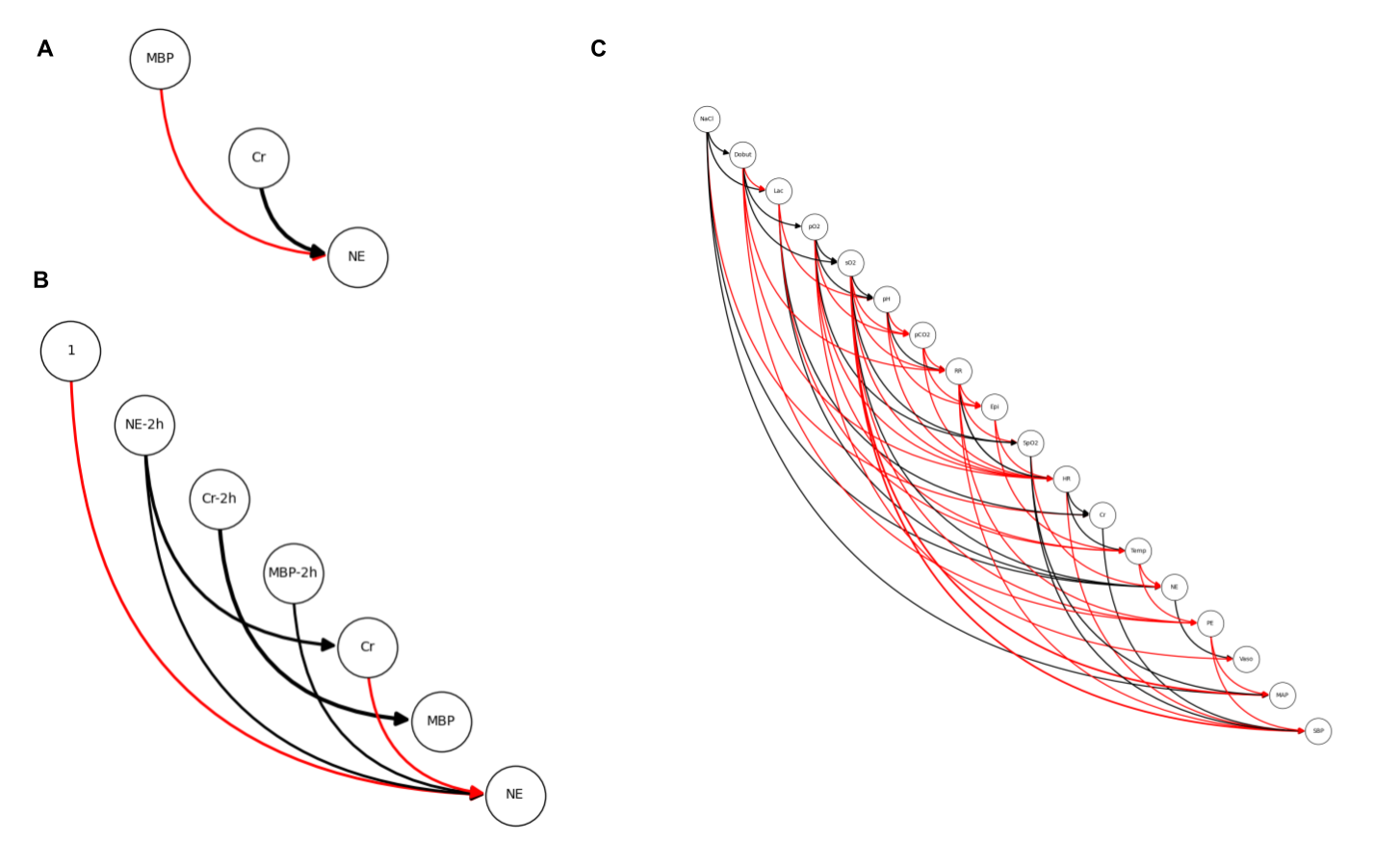}
\caption{Learned ordered inter-task dependency graphs.\\
\footnotesize
Graphical representations of the models reported in
Table~\ref{tab:shorterm}.
\textbf{A}: StructGP.
\textbf{B}: StructGP with 2-hour lag features.
\textbf{C}: StructGP with 15 additional tasks (see Additional File~1,
Section~S1.10).
Each directed edge represents a retained nonzero inter-task convolution
filter. Black and red edges indicate positive and negative peak filter
amplitudes, respectively, and edge thickness represents their absolute
magnitude. The graphs encode model-based ordered dependencies.
}
\label{fig:graph}
\end{figure}

In Figure~\ref{fig:graph}, we illustrate the learned dependency structures for three StructGP variants reported in Table~\ref{tab:shorterm}.
In panel A, a red link between MBP and NE, indicate that an increase in MBP is associated with a decrease in NE, whereas a black link between Cr and NE indicates that an increase in Cr is associated with a increase in NE.
In panel~B, a model with additional feature tasks, where “1” denotes a constant (intercept) pseudo-task, and “NE–2h” represents the 2-hour lagged version of NE.
Here, NE–2h is associated with increases in both Cr and NE.

\subsubsection{Long-term forecast}

\begin{table}[hbt!]
    \centering
\resizebox{\textwidth}{!}{
\begin{tabular}{llllll}
\toprule
 & \textbf{Models} & \multicolumn{4}{c}{\textbf{Output Tasks}} \\
 &   & \textbf{Average} & \textbf{NE} & \textbf{Cr} & \textbf{MBP} \\
 \midrule
 &  & \multicolumn{4}{c}{ \textbf{RMSE [95\% CI]}} \\
 \multicolumn{2}{l}{\textbf{Input Tasks: NE, Cr, MBP}} \\
 & StructGP & 1.04 [1.00, 1.07] & 0.97 [0.94, 1.00] & 1.17 [1.08, 1.25] & 0.97 [0.96, 0.98] \\
 &  LP-(fixed)StructGP &  1.01 [0.98, 1.03] & 0.93 [0.90, 0.97] & 1.10 [1.03, 1.17] & 1.00 [0.99, 1.01]   \\ 
 & LP-StructGP & 0.97 [0.94, 1.00] & 0.96 [0.93, 1.00] & 0.95 [0.88, 1.03] & 0.99 [0.98, 1.01] \\
& GraFITi & 0.95 [0.92, 0.98] & 0.89 [0.85, 0.92] & 1.00 [0.93, 1.08] & 0.97 [0.96, 0.98] \\
\midrule
 &  & \multicolumn{4}{c}{ \textbf{Coverage [95\% CI]}} \\
 \multicolumn{2}{l}{\textbf{Input Tasks: NE, Cr, MBP}} \\
 & StructGP & 0.91 [0.91, 0.92] & 0.89 [0.88, 0.90] & 0.90 [0.89, 0.92] & 0.95 [0.94, 0.95] \\
 & LP-(fixed)StructGP & 0.87 [0.87, 0.88] & 0.88 [0.87, 0.89] & 0.85 [0.83, 0.87] & 0.90 [0.89, 0.90] \\
 & LP-StructGP & 0.93 [0.93, 0.94] & 0.91 [0.90, 0.92] & 0.94 [0.92, 0.95] & 0.96 [0.95, 0.96] \\
% \midrule
\bottomrule
\end{tabular}

}
% \captionsetup{width=0.99\linewidth}
\caption{Long-term (six-day) horizon forecasting
\\\footnotesize
Root mean squared error (RMSE) and predictive coverage with bootstrap 95\% confidence intervals. Metrics are reported for up to a six-day forecasting horizon for Norepinephrine (NE), Creatinine (Cr), and mean blood pressure (MBP), after the first 24 hours of the ICU stay.
}
\label{tab:longterm}
\end{table}

Table~\ref{tab:longterm} compares StructGP with its latent-pathway extensions, LP-(fixed)StructGP and LP-StructGP, using three latent pathways.
In LP-(fixed)StructGP, the convolution filter parameters $\mathbf{H}$ were kept fixed to those learned by StructGP for computational efficiency.

Compared with StructGP, LP-StructGP achieved the best long-term performance, improving forecasts for creatinine (RMSE 0.95 [0.88–1.03] vs. 1.17 [1.08–1.25]).
Predictive calibration also improved, with coverage increasing from 0.91 [0.91–0.92] to 0.93 [0.93, 0.94].
LP-StructGP is competitive with GraFITi on Cr and MBP, but underperforms on NE (0.96 [0.93, 1.00] vs. 0.89 [0.85, 0.92]).

\begin{figure}[hbt!]%% placement specifier
\centering%% For centre alignment of image.
\includegraphics[width=\textwidth]{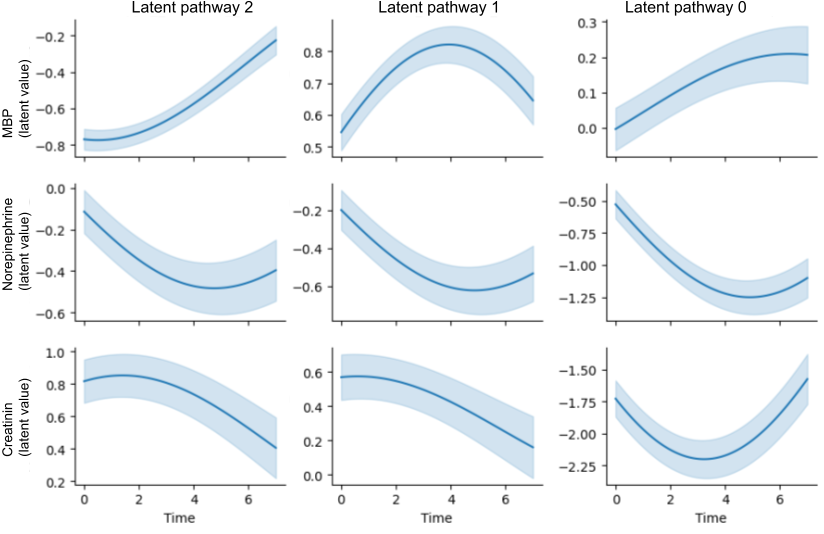}
\caption{Learned latent patient pathways visualization
\footnotesize
Learned from the LP-StructGP model reported in Table~\ref{tab:longterm}.
Each trajectory represents one inferred latent pathway across the 7-day ICU stay.
}
\label{fig:latent_pathways}
\end{figure}

Visualization of the learned latent pathways (Figure~\ref{fig:latent_pathways}) revealed two broad temporal patterns, primarily differing in their creatinine trajectories.
Qualitatively, these latent pathways appear to improve long-term forecasts for certain patients, as illustrated for two representative cases in Figure~\ref{fig:forecasts}.
A quantitative description of these pathways in the training cohort is provided in Supplementary Table~S4, where pathway membership probabilities below 0.5 were classified as \emph{unknown}.
This description reveals distinct characteristics across the learned pathways.
Compared with the other pathways, patients assigned to latent pathway~2 were older (median [Q1, Q3], 72 [61--80] years vs.\ 67 [56--77] years in the unknown group, 66 [56--76] years in pathway~0, and 63 [53--74] years in pathway~1; $p=0.002$), had a higher Charlson Comorbidity Index (6.0 [4.0--7.5] vs.\ 5.0 [3.0--7.0], 5.0 [2.0--6.0], and 5.0 [3.0--6.0], respectively; $p<0.001$), and experienced higher in-hospital mortality (47.2\% vs.\ 29.8\%, 29.9\%, and 27.7\%, respectively; $p<0.001$).

\begin{figure}[hbt!]%% placement specifier
\centering%% For centre alignment of image.
\includegraphics[width=\textwidth]{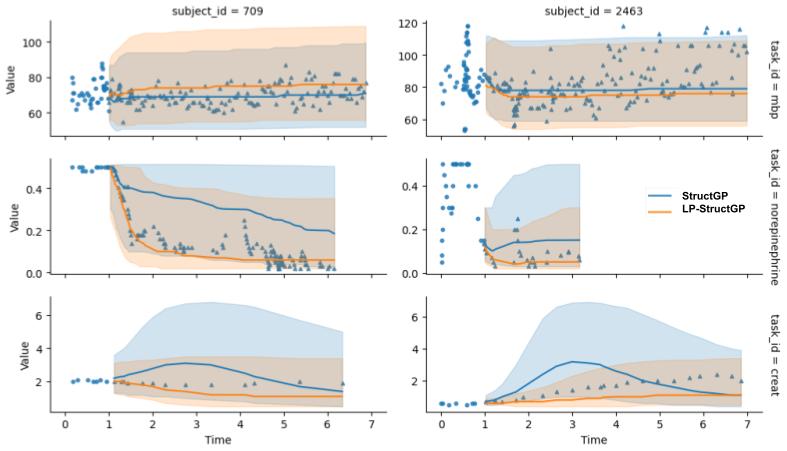}
  \caption{Comparison of six-day forecasts from StructGP and LP-StructGP.\\
\footnotesize
Subject-level forecasts for two selected patients, conditioned on
observations from the first 24~hours of the ICU stay. Observed values are shown
as points. Solid lines show the inverse-transformed posterior medians, and
shaded bands show pointwise 95\% posterior predictive intervals. The interval
bounds are obtained by inverse-transforming the 2.5th and 97.5th percentiles of
the Gaussian posterior predictive distribution in the transformed space.
}
\label{fig:forecasts}
\end{figure}

\subsection{PhysioNet Challenge dataset}\label{res:physionet}

\begin{table}[hbt!]
    \centering
\begin{tabular}{lll}
\toprule
 & \textbf{MAE $\times 10^{-2}$ [95\% CI]} & \textbf{MSE $\times 10^{-3}$ [95\% CI]} \\
\midrule
GraFITi (SOTA) & 3.50 [3.39, 3.62]  & 4.79 [4.27, 5.58]\\
StructGP & 4.06 [3.92, 4.21] & 7.55 [6.67, 8.48] \\
StructGP $+$ 3lat + time feat. & 3.72 [3.59, 3.86]  & 6.74 [6.05, 7.63] \\
\bottomrule
\end{tabular}
\caption{Comparison with SOTA on the PhysioNet challenge
\\\footnotesize
Mean absolute error (MAE) and Mean squared error (MSE) with bootstrap 95\% confidence intervals. Metrics are reported for up to a 1-day forecasting horizon macro-average for 41 tasks, after the first 24 hours of the ICU stay.
}
\label{tab:physionet}
\end{table}
The dataset comprises 41 time series from a total of 12,000 subjects (7,200 for training, 2,400 for development, and 2,400 for testing), with approximately 5.26 million observations in total (3,149,079 for training, 1,054,084 for development, and 1,056,460 for testing).
Table~\ref{tab:physionet} compares StructGP with the a strong published model on this dataset, \textbf{GraFITi}~\cite{yalavarthi2024grafiti}.  
The best-performing StructGP variant---incorporating time-feature pseudo-tasks and additional flexibility via three latent white-noise processes per task with a nested graphical structure---achieved an average MAE of $3.72~[3.59$--$3.86] \times 10^{-2}$, compared to $3.50~[3.39$--$3.62] \times 10^{-2}$ for GraFITi.  
However, the gap was larger in MSE, with StructGP reaching $6.74~[6.05$--$7.63] \times 10^{-3}$ versus $4.79~[4.27$--$5.58] \times 10^{-3}$.  
A detailed breakdown of per-variable performance is provided in Supplementary Table S1, listing the 18 tasks where StructGP underperformed, and Supplementary Table S2, which reports the 19 tasks where StructGP outperformed GraFITi.

For context, all other models evaluated by Yalavarthi \textit{et~al.}~\cite{yalavarthi2024grafiti}—except Zhang \textit{et~al.}~\cite{zhang2024irregular} (MAE $= 3.72 \times 10^{-2}$)—performed worse than both GraFITi and StructGP.
This includes NeuralFlow~\cite{bilovs2021neural} (MAE $= 4.67 \times 10^{-2}$), Latent-ODE~\cite{rubanova2019latent} ($4.23 \times 10^{-2}$), and CrossFormer~\cite{zhang2023crossformer} ($4.81 \times 10^{-2}$)
\footnote{The complete benchmark results are available at \url{https://github.com/yalavarthivk/GraFITi/tree/main}.}.
Moreover, when restricting evaluation to the top tercile of predictions with highest confidence (lowest posterior variance), StructGP slightly outperformed GraFITi, with an MAE of $2.46~[2.34$--$2.60] \times 10^{-2}$ versus $2.53~[2.44$--$2.64] \times 10^{-2}$ (Supplementary Table S3).

\section{Discussion}\label{sec:discussion}
We developed StructGP, a process-convolution model with an inter-task differentiable graphical structure, and LP-StructGP, an extension that embeds inter-subject latent pathways directly within the covariance. StructGP enables learning a shared, ordered dependency structure across time series in a differentiable manner, enhancing forecasting performance as the number of time series increases.
LP-StructGP learns latent trajectory motifs shared across patient subgroups, yielding more accurate long-horizon forecasts.
We further establish population identifiability of the ordered StructGP graph from the marginal likelihood under the stated assumptions. 
For LP-StructGP, the shared ordered inter-task graph is identifiable when the subject-level pathway filters are held fixed.
The simulations provide evidence that the learning procedure can
recover the inter-task graph and latent-pathway assignments under correct model
specification and sufficiently large cohorts. Recovery deteriorated when
uniform sampling was replaced by empirical MIMIC-IV-derived observation
schedules, illustrating the effect of sparse and uneven measurement designs.
On a MIMIC-IV septic shock cohort, incorporating structure became increasingly valuable with higher task dimensionality, improving both error and calibration; over longer horizons, latent pathways further enhanced creatinine forecasts and global coverage.
On the PhysioNet Challenge, StructGP achieved accuracy close to the current state of the art.
Together, these contributions support interpretable and transparent transfer across tasks and patients, advancing an uncertainty-aware, self-supervised structure learning framework for medium-sized, noisy clinical datasets.

In Experiment A, graph recovery approached perfect recovery at approximately
1{,}000 subjects, including when latent pathways were present. This experiment
was intentionally conducted under controlled conditions to isolate structural
recoverability: the simulated graphs were relatively small (10 nodes) and
sparse (mean degree of~2).
Importantly, the models were internally standardized, removing raw marginal
scale as a direct cue by normalizing each latent task component to unit
variance while preserving task-specific signal-to-noise ratios through the
rescaled shared observation noise. For StructGP, this construction corresponds
to the setting analyzed in Additional file~1, Section~S3.2. The simulation
results complement that population result by providing empirical evidence of
finite-sample graph recovery by the implemented procedure under controlled,
correctly specified conditions. For LP-StructGP, the simulations
demonstrate empirical joint recovery of graph and pathway assignments, whereas
the result in Additional file~1, Section~S3.3 establishes graph identifiability
only conditionally on fixed subject-level pathway filters.

With three tasks, the No Structure baseline contains only six possible directed
off-diagonal filters and performed similarly to StructGP, indicating little
predictive benefit from structural regularization in this low-dimensional
setting. With 18 tasks, however, the number of possible off-diagonal filters
increases to 306, together with their associated kernel parameters. The
sparsity and acyclicity constraints reduce this effective parameter space,
which can improve generalization and stabilize estimation. The marked
deterioration of the unconstrained baseline should therefore be interpreted
primarily as evidence of the benefit of structural regularization in the
higher-dimensional setting.

In Figure~\ref{fig:graph}, the model identified an association in which an increase in mean blood pressure (MBP) was linked to a decrease in norepinephrine (NE), whereas physiologically NE increases MBP.
This apparent reversal is in fact consistent with the difference in timescales: NE’s hemodynamic effect occurs over minutes~\cite{wiggins2012emergency}, whereas our dataset is sampled hourly, capturing instead the clinical adjustment of NE dosage in response to changes in MBP.
Together, these findings highlight the importance of allowing models to represent different directional effects at different temporal resolutions or time shifts.
A simple strategy is to include lagged “feature tasks,” as illustrated in Panel~B of Figure~\ref{fig:graph}, where a two-hour lag of NE (NE–2h) leads to an increase in creatinine—consistent with the known renal side effects of NE—while a contemporaneous link shows that increases in creatinine are associated with decreases in NE, potentially reflecting real corrective clinical interventions.

For long-term forecasts, LP-StructGP outperformed StructGP on the creatinine (Cr) task (Table~\ref{tab:longterm}).
It also revealed latent pathways characterized by increasing mean blood pressure (MBP), concurrent decreases in NE, and variable Cr trajectories—either gradually decreasing or decreasing followed by a late increase (Figure~\ref{fig:latent_pathways}), consistent with different patterns of septic shock recovery.
Exploratory comparisons in Supplementary Table~S4 showed that pathway
assignments were associated with several baseline characteristics and outcomes,
including chronic kidney disease, comorbidity burden, and in-hospital
mortality. For example, chronic kidney disease was less prevalent in
pathway~0 than in pathways~1 and~2 (0.9\% vs.\ 17.7\% and 19.5\%,
respectively; $p<0.001$).
These associations provide descriptive clinical
context for the learned trajectories, but do not establish the pathways as
validated septic-shock phenotypes or clinically actionable subgroups.
Interestingly, the task dependency structure was shared across latent pathways, preserving interpretability within this more flexible model.

\subsection{Related Work}

Compared with leading graph neural network models such as GraFITi~\cite{yalavarthi2024grafiti} on PhysioNet, the best StructGP variant underperformed on 18 tasks and outperformed on 19 tasks based on the MAE metric.
In addition, its predictions provide calibrated uncertainty estimates and remain interpretable, confirming process convolution models as a promising direction for continuous-time modeling of clinical time series.

From a scalability perspective, the current implementation is tailored for large cohorts with many patients but relatively few observations per patient.
Under this regime, StructGP successfully fit 5.26 million observations from the PhysioNet dataset (41 tasks) in approximately 45~minutes on a single NVIDIA A40 GPU, including 100 grid-search iterations and repeated minimization of the NMLL to satisfy the acyclicity constraint at each step (Additional file 1, Section S1.5).
Future work could extend scalability to settings with longer individual trajectories or larger numbers of tasks while remaining within an exact GP framework, by leveraging dedicated block-sparse routines for Gaussian process solvers~\cite{charlier2021kernel}.

From a structure learning standpoint, StructGP can be viewed as a continuous-time extension of DYNOTEARS~\cite{pamfil2020dynotears}, a necessary step for modeling clinical time series.
Dallakyan \textit{et al.} \cite{dallakyan2023learning} proposed a continuous-time structure learning approach based on a Gaussian linear additive SCM in the Fourier domain. Unlike their method, which estimates separate weight matrices at each frequency, StructGP assumes a single invariant structure across frequencies.

Gaussian processes (GPs) have long been applied to clinical time series.
Early work by Durichen \textit{et al.}~\cite{durichen2014multitask} introduced process convolution kernels with time-shift parameters, and later studies addressed population- versus patient-level modeling~\cite{liu2016learning, cheng2020sparse, cui2021hierarchical, karch2020gaussian}.
These approaches typically separate inter-task and inter-patient dependencies or rely on fixed hierarchical structures.
In contrast, StructGP learns inter-task relationships within subjects, whereas
LP-StructGP additionally learns between-subject latent-pathway structure,
through covariance models trained using marginal-likelihood-based objectives.

\section{Limitations and perspectives}
The main limitation of the current implementation is that, since this study
focused on developing the model's structural assumptions, we restricted our
experiments to Gaussian likelihoods.
Although the data were standardized to roughly satisfy the Gaussian
assumption, more flexible likelihood formulations
\cite{chan2011generalized, bonilla2016generic, murray2018mixed,
ramchandran2021latent} would be necessary to capture the heterogeneity of
clinical time series and to jointly model discrete patient states, such as
intubation status, missingness indicators, or mortality.
Gaussian observational laws do not generically identify an ordered graph
without structural restrictions~\cite[page 138]{peters2017elements}. Our
result is specific to the anchored acyclic StructGP class and requires a rich
observation design and identifiable noise; after internal standardization, it
also requires shared raw noise and separation of the nugget.
More broadly, continuous variables represent only a subset of EHR data \cite{neuraz2020natural, escudie2017novel}, and
incorporating unstructured modalities such as clinical text and imaging
 would require dedicated multimodal extensions.

The simulation study was designed to isolate structural recovery under
controlled, correctly specified conditions. Experiment B additionally
evaluates uneven measurement densities and short observation sequences using
empirical MIMIC-IV schedules, while Experiment C examines selection among
incorrect candidate numbers of latent pathways. However, the schedules in
Experiment B were sampled independently of the simulated trajectories and
therefore do not reproduce informative observation processes. Moreover,
excluding patients below the 20th percentile of measurement availability in
the clinical evaluation limits conclusions for very sparse histories. When
measurement decisions depend on unobserved clinical states, future extensions
should jointly model the clinical and observation processes, for example
through temporal point-process or non-Gaussian measurement-indicator
likelihoods.

The learned graph should be interpreted as a model-based ordered dependency
structure, rather than as a causal graph. Under the stationary Gaussian
assumptions, its sparsity pattern encodes ordered conditional dependencies
between process components. In observational EHR data, however, these
dependencies may reflect physiology, treatment and measurement policies,
sampling resolution, or clinician responses. A causal interpretation would
require additional assumptions concerning confounding, interventions, and the
observation process that are not established in the present study.

From a design standpoint, LP-StructGP currently faces scalability limitations.
In practice, one must choose either a small number of latent pathways or a small number of tasks, since the feature dimension of the inter-subject component scales as $m \times k \times p$, where $m$ is the number of basis functions in the HSGP approximation, $k$ the number of tasks, and $p$ the number of latent pathways.
A simple remedy is to assign latent pathway components only to a subset of nodes (e.g., root nodes).
An even more appealing direction would be to restrict latent pathways to latent unobserved tasks, although inference and identifiability in that setting would be more challenging.

The group-level component of the model provides greater flexibility than the individual component in extrapolation settings, as it interpolates time series across training patients to identify shared patterns and trends.
However, it remains limited to patients following a single latent pathway within the observed time period. Introducing additional patient-specific parameters at selected time points could enable transitions between different dynamical motifs.

Further experiments are needed to fully characterize the model’s properties, including its causal properties,  transfer learning capabilities, and comparative performance across heterogeneous datasets. Nevertheless, this study introduces, two continuous-time models that leverage latent structure in clinical data to improve forecasting, paving the way toward algorithmic approaches that bridge the gap from real-life data to knowledge.

\section{Conclusion}

We developed StructGP, a process-convolution model that learns dependency structures among irregularly sampled time series, and its extension LP-StructGP, which captures dependencies across subjects through shared latent pathways. We showed that incorporating these structures in continuous-time models improves forecasting of real-world clinical time series while maintaining interpretability and calibrated uncertainty estimates.
Future work should further characterize the conditions under which a causal interpretation is valid, paving the way toward a self-supervised framework of structured random functions that algorithmically bridges the gap from natural data streams to knowledge.

\backmatter

\bmhead{Acknowledgements}

We would like to thank INRIA, INSERM, and AP-HP for their administrative, technical, and material support.
We thank Alexander Reisach for his interest in this work and for helpful feedback.

\section*{Declarations}\label{sec:declaration}

\subsection*{Ethics}\label{sec:ethics}
Permission to use the data was obtained for MIMIC-IV (No. 45398938).
Because of the de-identified nature of the data, informed consent was waived.
Consent was obtained for the original data collection.

\subsection*{Data availability}\label{sec:data_availability}

This study was performed with the data from the MIMIC-IV version 1.0.
Even though datasets are de-identified, restrictions have been imposed on data sharing since they contain sensitive information.
Conventions are signed by researchers before any access to the data.
For data access, interested researchers must fulfill all of the following requirements: be a credentialed user of \href{physionet}{https://physionet.org/}, finish required training, and sign the data use agreement for the project.

Reproducing the PhysioNet Challenge experiment with GraFIti model can be made easily at \href{github}{https://github.com/yalavarthivk/GraFITi}.

\subsection*{Declaration of generative AI and AI-assisted technologies in the manuscript preparation process.
}\label{sec:ai_declar}
Statement: During the preparation of this work the authors used GPT-5 (OpenAI) in order to proofread and reformulate sentences, paragraphs, or math formulas in latex.
GPT-5 was also used as an assistant for the analysis of model properties and for proof engineering, including the structuring, checking, and refinement of mathematical arguments.
After using this tool, the authors reviewed and edited the content as needed and take full responsibility for the content of the published article.

\subsection*{Contributions}
IL had full access to all of the data in the study and takes responsibility for the integrity of the data and the accuracy of the data analysis.
Concept and design: IL, FB.
Acquisition, analysis, or interpretation of data: AK, IL, AB, JF, FB.
Drafting of the manuscript: IL.
Critical revision of the manuscript for important intellectual content: IL, FB, JF, AB.
Administrative, technical, or material support: AB.
Supervision: FB, AB.

\subsection*{Funding}
This work was supported in part by the French government under the management of ANR as part of the “Investissements d’avenir” program, reference ANR-19-P3IA-0001 (PRAIRIE 3IA Institute).
 
\subsection*{Declaration of interest}
The authors declare that they have no known competing financial interests or personal relationships that could have appeared to influence the work reported in this paper.

\subsection*{Additional file}

Additional file 1: Supplementary methods, supplementary results, supplementary figures, and supplementary tables.
PDF file containing detailed mathematical derivations, algorithmic details, additional experimental details, Supplementary Figures S1--S3, and Supplementary Tables S1--S4.

\clearpage
\bibliography{structgp}% common bib file

\end{document}

\section{Supplementary methods}

\subsection{Detailed StructGP covariance}
\label{app:inter_task}

Let \(\mathbf S^{\mathrm{off}}\) denote the matrix of learned inter-task
coefficients, with zero diagonal. The StructGP filter is
\[
    \mathbf H(t)
    =
    \bigl(\mathbf I+\mathbf S^{\mathrm{off}}\bigr)
    \circ\mathbf L(t).
\]
For notational simplicity, throughout the appendix we denote the full
amplitude matrix \(\mathbf I+\mathbf S^{\mathrm{off}}\) by \(\mathbf S\).
Thus,
\[
    \mathbf H(t)=\mathbf S\circ\mathbf L(t),
    \qquad
    S_{vv}=1.
\]
Only the off-diagonal entries of \(\mathbf S\) are learned graph
coefficients; acyclicity constraints and sparsity penalties apply only to
these entries.

Each task \( Y_v(t) \) then results from the convolution of independent latent white noise processes \( \{ w_u(t) \}_{u=1}^k \):
\begin{align}
    Y_v(t) &= \sum_{u=1}^k (H_{vu} * w_u)(t),
\end{align}
so that the covariance between tasks \( v \) and \( v' \) is given by
\begin{align}
    k_{vv'}(t, t') &= \sum_{u=1}^k (H_{vu} * H_{v'u})(t - t').
\end{align}

Since the white noise processes are mutually independent, only shared sources $u$ contribute to the covariance.
We assume here that all filters $H_{vu}(t)$ are Gaussian functions of the form
\begin{align}
    H_{vu}(t) = S_{vu} \exp\!\left(-\frac{t^2}{\ell_{vu}}\right),
\end{align}
where $S_{vu}$ encodes the connectivity strength from task $u$ to task $v$, and $\ell_{vu}>0$ controls its temporal smoothness.  
Under this assumption, each convolution term has a closed-form expression:
\begin{align}
    (H_{vu} * H_{v'u})(t - t') 
    = S_{vu} S_{v'u} \sqrt{\pi\frac{\ell_{vu}\ell_{v'u}}{\ell_{vu} + \ell_{v'u}}}
      \exp\!\left(-\frac{(t - t')^2}{\ell_{vu} + \ell_{v'u}}\right).
\end{align}

The overall covariance between tasks \(v\) and \(v'\) is obtained by summing
over the latent sources that enter both tasks. To include the fixed
self-source of each task, define the augmented parent set
\[
    \mathrm{Pa}^{+}(v)
    :=
    \mathrm{Pa}(v)\cup\{v\}.
\]
Then
\begin{align}
    k_{vv'}(t,t')
    &=
    \sum_{u\in
        \mathrm{Pa}^{+}(v)\cap\mathrm{Pa}^{+}(v')}
    (H_{vu}*H_{v'u})(t-t').
\end{align}
Here \(\mathrm{Pa}(v)\) denotes the set of off-diagonal parent tasks of
\(v\) in the DAG \(\mathcal G\), while \(v\in\mathrm{Pa}^{+}(v)\)
represents its fixed identity self-connection.

The learnable kernel parameters are grouped into $\theta = \{\mathbf{S}, \mathbf{L}\}$, with $\mathbf{S} \in \mathbb{R}^{k \times k}$ encoding the graph structure and $\mathbf{L} \in \mathbb{R}_+^{k \times k}$ the temporal scales.

We also include an observation-level noise term in the kernel.
In the SNR-preserving internally standardized model analyzed in Section~\ref{app:snr_preserving_identifiability}, this noise term is obtained by rescaling a shared raw observation-noise variance with the same taskwise normalization used for the latent process-convolution component.
Variants with task-specific noise terms are also considered, but these relaxed models are not covered by the SNR-preserving identifiability result.

\subsection{Internal standardization}\label{app:internal_standardization}

The internal standardization used in StructGP is inspired by the internally
standardized structural causal models of Ormaniec et
al.~\cite{ormaniec2024standardizing}, but adapted to the
process-convolution setting.
In iSCMs, the standardized variables are the observed variables.
In StructGP, we instead standardize the latent process-convolution component of
each task, so that each latent task has unit marginal variance before
observation noise is added.
This normalization reduces the dependence of the learned amplitudes \(S_{vu}\)
on arbitrary marginal task scales, so that they mainly encode the relative
strength of dependencies between tasks.

For Gaussian filters of the form
\begin{align}
    H_{vu}(t)
    =
    S_{vu}
    \exp\!\left(-\frac{t^2}{\ell_{vu}}\right),
\end{align}
the marginal variance of the latent process-convolution component of task \(v\)
is obtained by integrating the squared filters over time:
\begin{align}
    q_v
    &:=
    \sum_{u=1}^k
    \int_{-\infty}^{\infty}
    H_{vu}^2(t)\,dt  \nonumber \\
    &=
    \sum_{u=1}^k
    S_{vu}^2
    \sqrt{\frac{\pi \ell_{vu}}{2}} .
\end{align}
We define the normalization factor for each task \(v\) as
\begin{align}
    s_v
    =
    \sqrt{q_v}
    =
    \left(
        \sum_{u=1}^k
        S_{vu}^2
        \sqrt{\frac{\pi \ell_{vu}}{2}}
    \right)^{1/2}.
\end{align}
The standardized amplitudes are then given by
\begin{align}
    \widetilde S_{vu}
    =
    \frac{S_{vu}}{s_v},
\end{align}
and the corresponding normalized filters are
\begin{align}
    \widetilde H_{vu}(t)
    =
    \widetilde S_{vu}
    \exp\!\left(-\frac{t^2}{\ell_{vu}}\right).
\end{align}
By construction, these normalized filters give unit marginal variance to the
latent process-convolution component of each task:
\begin{align}
    \sum_{u=1}^k
    \int_{-\infty}^{\infty}
    \widetilde H_{vu}^2(t)\,dt
    =
    1 .
\end{align}
Since \(s_v>0\), this row-wise rescaling does not change the support of the
convolution filter. Therefore, it does not change the ordered-dependence graph
encoded by \(H\) or \(S\).

By default, the internally standardized StructGP model uses a
signal-to-noise-ratio-preserving construction. The taskwise scale used to
normalize the latent process-convolution component is also applied to the
shared raw observation-noise term.
Thus, if the raw observation noise has variance \(\sigma^2\), the standardized
nugget for task \(v\) becomes
\begin{align}
    \widetilde\sigma_v^2
    =
    \frac{\sigma^2}{q_v}.
\end{align}
Since the standardized latent variance is one, the standardized
signal-to-noise ratio is
\begin{align}
    \operatorname{SNR}_v^{\mathrm{std}}
    =
    \frac{1}{\widetilde\sigma_v^2}
    =
    \frac{q_v}{\sigma^2}
    =
    \operatorname{SNR}_v^{\mathrm{raw}} .
\end{align}
This is the version used in the graph-recovery simulations and analyzed in
Section~\ref{app:snr_preserving_identifiability}. It preserves the relative
noise information needed for graph identifiability while still normalizing the
latent process scale.

\subsection{Detailed LP-StructGP formulation}
\label{app:latent_pathway}

To capture shared temporal dynamics across subjects, we extend StructGP with an
additional latent layer representing group-level trajectories, referred to as
\textit{latent pathways}.
We introduce \(p\) mutually independent \(k\)-dimensional white-noise processes
\(\{\mathbf w_u(t)\}_{u=1}^p\), each giving rise to a multivariate latent
process \(\mathbf Q_u(t)\in\mathbb R^k\) through the shared inter-task filter
\(\mathbf H\):
\begin{align}
    \mathbf Q_u(t)
    &=
    (\mathbf H*\mathbf w_u)(t).
\end{align}
All latent pathways therefore share the same inter-task dependency structure,
consistent with a \textit{biological invariance hypothesis} assuming common
task-dependency mechanisms across subjects.

For each subject \(i\), the pathway weights are obtained by applying a softmax
transformation to unconstrained subject-specific logits:
\begin{align}
    \boldsymbol{\pi}_i
    &=
    \operatorname{softmax}
    \!\left(\mathbf S^{(\mathrm{sub})}_i\right), \\
    \pi_{iu}
    &=
    \frac{
        \exp\!\left(S^{(\mathrm{sub})}_{iu}\right)
    }{
        \sum_{a=1}^{p}
        \exp\!\left(S^{(\mathrm{sub})}_{ia}\right)
    }.
\end{align}
Thus, \(\pi_{iu}\in(0,1)\) and
\(\sum_{u=1}^{p}\pi_{iu}=1\).
These normalized weights provide a differentiable soft gating mechanism
\cite{jacobs1991adaptive, shazeer2017outrageously}; the largest weight
identifies the predominant pathway for a subject while allowing contributions
from multiple pathways.

The subject--pathway coupling filter is separable across tasks and is defined
as
\begin{align}
    \mathbf G_{iu}(s)
    &=
    g_{iu}(s)\mathbf I_k, \\
    g_{iu}(s)
    &=
    \pi_{iu}
    \exp\!\left(
        -\frac{(s-\tau_{iu})^2}
        {\ell^{(\mathrm{sub})}_{iu}}
    \right),
\end{align}
where \(\ell^{(\mathrm{sub})}_{iu}>0\) controls temporal smoothness and
\(\tau_{iu}\) represents a subject-specific temporal shift along the shared
latent pathway. The scalar filter \(g_{iu}\) therefore acts identically on all
task components of \(\mathbf Q_u\).

Each subject \(i\) expresses an individualized trajectory as a mixture of the
latent pathways combined with a subject-specific residual component:
\begin{align}
    \mathbf Y_i(t)
    &=
    \sum_{u=1}^{p}
    (\mathbf G_{iu}*\mathbf Q_u)(t)
    +
    (\mathbf H*\mathbf w_i)(t),
\end{align}
where the subject-specific white-noise processes \(\mathbf w_i\) are mutually
independent and independent of all pathway processes \(\mathbf w_u\).
Because \(\mathbf G_{iu}(s)=g_{iu}(s)\mathbf I_k\), the first convolution may
equivalently be written as
\[
    (\mathbf G_{iu}*\mathbf Q_u)(t)
    =
    (g_{iu}*\mathbf Q_u)(t),
\]
with the scalar convolution applied componentwise.
Let \(r=t-t'\) and
\[
    \mathbf K_H(r)
    =
    (\mathbf H*\mathbf H^\top)(r)
\]
denote the shared inter-task covariance. For pathway \(u\), define the
cross-correlation between the subject-specific coupling filters as
\begin{align}
    R_{ii'u}(r)
    &:=
    (g_{iu}\star g_{i'u})(r) \\
    &:=
    \int_{-\infty}^{\infty}
    g_{iu}(s)g_{i'u}(s-r)\,ds.
\end{align}
The covariance between subjects \(i\) and \(i'\) is then
\begin{align}
    \operatorname{Cov}
    \!\left[
        \mathbf Y_i(t),
        \mathbf Y_{i'}(t')
    \right]
    &=
    \delta_{ii'}\mathbf K_H(r)
    +
    \sum_{u=1}^{p}
    \left[
        \mathbf K_H*R_{ii'u}
    \right](r),
\end{align}
where the scalar convolution with \(R_{ii'u}\) is applied entrywise to
\(\mathbf K_H\).

For the shifted Gaussian filters defined above,
\begin{align}
    R_{ii'u}(r)
    &=
    \pi_{iu}\pi_{i'u}
    \sqrt{
        \pi
        \frac{
            \ell^{(\mathrm{sub})}_{iu}
            \ell^{(\mathrm{sub})}_{i'u}
        }{
            \ell^{(\mathrm{sub})}_{iu}
            +
            \ell^{(\mathrm{sub})}_{i'u}
        }
    }
    \nonumber\\
    &\quad\times
    \exp\!\left[
        -
        \frac{
            \left(
                r-\tau_{iu}+\tau_{i'u}
            \right)^2
        }{
            \ell^{(\mathrm{sub})}_{iu}
            +
            \ell^{(\mathrm{sub})}_{i'u}
        }
    \right].
\end{align}
Thus, the cross-subject covariance depends on the relative temporal shift
\(\tau_{iu}-\tau_{i'u}\).

The complete parameter set of LP-StructGP is
\begin{align}
    \theta
    =
    \{
        \mathbf S,
        \mathbf L,
        \mathbf S^{(\mathrm{sub})},
        \mathbf L^{(\mathrm{sub})},
        \boldsymbol{\tau}^{(\mathrm{sub})}
    \},
\end{align}
where \((\mathbf S,\mathbf L)\) define the shared inter-task structure, while
\((\mathbf S^{(\mathrm{sub})},
\mathbf L^{(\mathrm{sub})},
\boldsymbol{\tau}^{(\mathrm{sub})})\) govern the subject--pathway
interactions. The weights \(\boldsymbol{\pi}\) are not listed separately
because they are deterministic softmax transformations of
\(\mathbf S^{(\mathrm{sub})}\).

In practice, the shared latent-pathway and subject-specific components are
combined using fixed coefficients controlled by a hyperparameter that balances
their relative contributions. These fixed coefficients are omitted from the
equations above because they only rescale the corresponding covariance
components and are not included in the learned parameter set \(\theta\).

\subsection{Gaussian process learning and inference}\label{app:gp_learning}

We consider a collection of $n$ scalar observations $y(\mathbf{x})$, drawn from $r$ individuals and $k$ tasks:
\[
\{y(\mathbf{x}) : \mathbf{x} = (i, j, t), \;
i \in \{1, \ldots, r\}, \;
j \in \{1, \ldots, k\}, \;
t \in \mathbb{R}^+\}.
\]
Each observation is indexed by the triplet $\mathbf{x} = (i, j, t)$, corresponding to the subject index, task index, and observation time, respectively.

We use a zero-mean prior after preprocessing, $\mathbb{E}[y(\mathbf{x})] = 0$, since the posterior mean will be data-dependent.  
The process is then fully characterized by its covariance function:
\[
y(\mathbf{x}) \sim \mathcal{GP}(0, k(\mathbf{x}, \mathbf{x}')),
\qquad
k(\mathbf{x}, \mathbf{x}') = \mathbb{E}[y(\mathbf{x}) y(\mathbf{x}')].
\]

Gaussian process regression consists of two main steps.  
First, we place a prior distribution on $y(\mathbf{x})$, with modeling assumptions encoded in the kernel $k$.  
The kernel hyperparameters $\theta$ (which include the convolution filter parameters and the noise variance $\sigma^2$) are learned by maximizing the log marginal likelihood of the training observations $\mathbf{y}$ given the inputs $\mathbf{X}$~\cite{mardia1984maximum}:
\begin{equation}
    \log p(\mathbf{y}|\mathbf{X}, \theta)
    = -\frac{1}{2}\mathbf{y}^{\top}(\mathbf{K} + \sigma^2 \mathbf{I})^{-1}\mathbf{y}
      -\frac{1}{2}\log|\mathbf{K} + \sigma^2 \mathbf{I}|
      -\frac{n}{2}\log(2\pi),
    \label{eq:gp_nmll}
\end{equation}
where $\mathbf{K}$ is the $n \times n$ covariance matrix with entries $\mathbf{K}_{uv} = k(\mathbf{x}_u, \mathbf{x}_v)$ for $\mathbf{x}_u, \mathbf{x}_v \in \mathbf{X}$.  
This objective balances data fit (the quadratic term) and model complexity (the log-determinant term), corresponding geometrically to finding the smallest covariance ellipsoid that adequately explains the data.

In the second step, we define the joint prior distribution between $n$ training observations $\mathbf{y}$ and $m$ test observations $\mathbf{y}_*$, indexed by
$\mathbf{X} \in (\mathbb{N}, \mathbb{N}, \mathbb{R}^+)^n$
and
$\mathbf{X}_* \in (\mathbb{N}, \mathbb{N}, \mathbb{R}^+)^m$, respectively:
\begin{equation}
    \begin{bmatrix}
    \mathbf{y} \\
    \mathbf{y}_*
    \end{bmatrix}
    \sim
    \mathcal{N}\!\left(
    \mathbf{0},
    \begin{bmatrix}
    \mathbf{K} + \sigma^2 \mathbf{I} & \mathbf{K}_* \\
    \mathbf{K}_*^{\top} & \mathbf{K}_{**} + \sigma^2 \mathbf{I}
    \end{bmatrix}
    \right),
    \label{eq:gp_joint}
\end{equation}
where $\mathbf{K}$, $\mathbf{K}_*$, and $\mathbf{K}_{**}$ denote the covariance matrices over training, cross, and test inputs, respectively.

Conditioning on $\mathbf{y}$ yields the posterior predictive distribution:
\begin{align}
    \mathbf{y}_* \mid \mathbf{y}, \mathbf{X}, \mathbf{X}_*
    &\sim \mathcal{N}(\boldsymbol{\mu}_*, \boldsymbol{\Sigma}_*), \\
    \text{where} \quad
    \boldsymbol{\mu}_* &= \mathbf{K}_*^{\top}(\mathbf{K} + \sigma^2 \mathbf{I})^{-1}\mathbf{y}, \\
    \boldsymbol{\Sigma}_*
    &= \mathbf K_{**}+\sigma^2\mathbf I
    - \mathbf K_*^{\top}
    (\mathbf K+\sigma^2\mathbf I)^{-1}\mathbf K_*.
\end{align}

In practice, the kernel $k_\theta(\mathbf{x}, \mathbf{x}')$ encodes the structured covariance induced by either StructGP or LP-StructGP, with shared parameters across tasks and patients.

\subsection{Differentiable structure learning: algorithmic details}\label{app:diff_structure_learning_details}

In StructGP, we impose that $\mathbf S^{\mathrm{off}}$ encodes the weighted adjacency matrix of a sparse directed acyclic graph (DAG) $\mathcal{G}$, i.e., that $\mathbf S^{\mathrm{off}}$ is sparse and lower-triangular up to a permutation of the task order.
To jointly learn the graph structure, the task ordering, and the sparsity pattern, we adapt the NOTEARS algorithm~\cite{zheng2018dags}.
NOTEARS introduces a differentiable characterization of acyclicity based on the trace of the matrix exponential of the adjacency matrix (see \ref{app:continuous_sl}).

Our goal is therefore to solve the following constrained optimization problem:
\begin{align}
  \theta^*
  &= \operatorname*{argmin}_\theta
     \Big[-\log \mathcal{L}(\mathbf{y}, \mathbf{X}, \theta)
     + \mathcal{P}_{\lambda}(\mathbf S^{\mathrm{off}})\Big] \\
  &\text{s.t.} \quad
  h(\mathbf S^{\mathrm{off}}) := \operatorname{Tr}\big(\exp(\mathbf S^{\mathrm{off}}\circ\mathbf S^{\mathrm{off}})\big) - k = 0, \nonumber
\end{align}
where $\mathcal{P}_{\lambda}$ is a sparsity-inducing penalty with strength $\lambda$,
 $\mathcal{L}$ is the process marginal likelihood and $\theta$ the set of learnable parameters.

This problem is solved using the augmented Lagrangian method~\cite{nemirovsky1999optimization}, which converts the constrained problem into a sequence of unconstrained subproblems (the \emph{primal}) and updates a dual variable by gradient ascent (see \ref{app:lagrang}).
The primal objective is the penalized likelihood augmented with a quadratic penalty term enforcing the acyclicity constraint.

To make $\mathcal{P}_{\lambda}$ differentiable everywhere, we use the SmoothL1-SC approximation to the $\ell_1$ norm~\cite{schmidt2009optimization}:
\[
\mathcal{P}_{\lambda}(\mathbf S^{\mathrm{off}})
= \lambda \sum_i \frac{1}{\beta_{\text{L1}}}
\Big[
\log\big(1 + e^{\beta_{\text{L1}} S_i}\big)
+ \log\big(1 + e^{-\beta_{\text{L1}} S_i}\big)
\Big],
\]
where a large $\beta_{\text{L1}}$ yields a close approximation to the standard $\ell_1$ penalty.

We use Adam~\cite{kingma2014adam} as the gradient-based optimizer.
The sparsity weight $\lambda_*$ is selected by grid search to minimize either an AIC-like criterion,
\[
\text{AIC} = 2\|\mathbf S^{\mathrm{off}}\|_0 - 2\log \mathcal{L}(\mathbf{y}, \mathbf{X}, \theta),
\]
or a validation loss.
The search is conducted on a logarithmic scale from $\lambda_{\max}$ to $\lambda_{\min}$ with warm starts.

Because solving the augmented Lagrangian problem to high precision is computationally expensive and may cause numerical instability when the penalty coefficient $\rho$ becomes large, we only solve it approximately, with a loose tolerance for the acyclicity constraint (typically $\epsilon = 0.01$, see \ref{app:lagrang}).
To guarantee acyclicity in the final graph, we apply a hard-thresholding step to $\mathbf S^{\mathrm{off}}$, removing all entries below the smallest value that ensures $h(\mathbf S^{\mathrm{off}}) = 0$.

All implementation details and code are available at
\href{https://gitlab.com/lerner.ivan/structgp_pub}{\underline{gitlab}}.

\subsubsection{The matrix exponential acyclicity constraint}\label{app:continuous_sl}

\begin{figure}[htp]
\centering
\begin{minipage}{0.30\textwidth}
\begin{tikzpicture}[node distance={20mm}, thick, main/.style = {draw, circle}] 
\node[main] (1) {$Y_0$}; 
\node[main] (2) [above right of=1] {$Y_1$}; 
\node[main] (3) [below right of=1] {$Y_2$}; 
\draw[->] (1) -- node[midway, above, sloped] {\small $b_{01}$} (2); 
\draw[->] (2) -- node[midway, above, sloped] {\small $b_{12}$} (3); 
\draw[->] (3) -- node[midway, above, sloped] {\small $b_{20}$} (1); 
\end{tikzpicture} 
\end{minipage}%
\hfill
\begin{minipage}{0.70\textwidth}
\small
$\textbf{B} =\  \left[\begin{matrix}0 & b_{01} & 0\\0 & 0 & b_{12}\\b_{20} & 0 & 0\end{matrix}\right]$\\
$\textbf{B}^2 = \left[\begin{matrix}0 & 0 & b_{01} b_{12}\\b_{12} b_{20} & 0 & 0\\0 & b_{01} b_{20} & 0\end{matrix}\right]$\\
$ \textbf{B}^3 = \left[\begin{matrix}b_{01} b_{12} b_{20} & 0 & 0\\0 & b_{01} b_{12} b_{20} & 0\\0 & 0 & b_{01} b_{12} b_{20}\end{matrix}\right]$
\end{minipage}%
\caption{Counting cycles in adjacency matrices using the matrix exponential.}
\label{fig:notears}
\end{figure}

The NOTEARS formulation expresses the acyclicity of a directed graph as a differentiable function of its weighted adjacency matrix~\cite{zheng2018dags}.
Let $\mathbf{B} \in \mathbb{R}^{k\times k}$ denote the weighted adjacency matrix of a graph $\mathcal{G}$ with nodes $\{Y_1,\dots,Y_k\}$.
An edge $Y_u \to Y_v$ exists if and only if $b_{uv} \neq 0$.
The graph $\mathcal{G}(\mathbf{B})$ is acyclic if and only if
\[
h(\mathbf{B}) := \operatorname{Tr}\!\big(\exp(\mathbf{B}\circ\mathbf{B})\big) - k = 0,
\]
where $\circ$ denotes the Hadamard (elementwise) product.
Expanding the matrix exponential,
\[
\operatorname{Tr}(\exp \mathbf{S})
= \operatorname{Tr}(I) + \operatorname{Tr}(\mathbf{S})
  + \tfrac{1}{2!}\operatorname{Tr}(\mathbf{S}^2) + \cdots,
\]
shows that $\operatorname{Tr}(\mathbf{S}^m)$ counts weighted cycles of length $m$  (see Figure \ref{fig:notears}).
Thus $h(\mathbf{B}) = 0$ ensures that all such cycles vanish, yielding a differentiable characterization of acyclicity.
For a detailed discussion, see~\cite{vowels2022d}.

\newpage
\subsubsection{The augmented Lagrangian optimization algorithm}\label{app:lagrang}

\begin{mdframed}[linewidth=1.5pt, % Adjust the thickness of the frame
    innerleftmargin=10pt, % Adjust the left margin
    innerrightmargin=10pt, % Adjust the right margin
    frametitle={}]
Given:
\begin{itemize}
    \item Objective function: $f(\theta) = -\log \mathcal{L}(\mathbf{y}, \mathbf{X}, \theta) + \mathcal{P}_{\lambda}(\mathbf{S})$
    \item Constraint: $g(\theta) = \operatorname{Tr}(\exp(\mathbf{S} \circ \mathbf{S} )) - k = 0$
    \item Convergence criteria: constraint tolerance $\epsilon$ and $\rho_{max}$
    \item Primal solver: \textit{Solver}
\end{itemize}
Define:
\begin{itemize}
    \item Lagrange multiplier $\alpha^{(k)}$ at step $k$
    \item Augmented Lagrangian: $L(\theta, \alpha^{(k)}, \rho) = f(\theta) + \alpha^{(k)} g(\theta) + \frac{\rho}{2}g(\theta)^2$
\end{itemize}
Do:
\begin{enumerate}
    \item Choose initial guess $\theta^{(0)}$, Lagrange multipliers $\alpha^{(0)}=0$ and $\rho = 1$
    \item For $k = 0, 1, 2, \ldots$
    \begin{enumerate}
        \item  While $\rho < \rho_{max}$ update $\theta^{(k+1)}$.
        \begin{itemize}
            \item  Minimize the augmented Lagrangian function:
            \[
            \theta^{(k+1)} = \textit{Solver}\big(L(\theta, \alpha^{(k)}, \rho)\big)
            \]
            \item Break if $g(.)$ sufficiently decreases:
            \[g(\theta^{(k+1)}) < 0.25\ g(\theta^{(k)}) \]
            \item Else augment $\rho$: 
            \[\rho = 10 \ \rho \]
        \end{itemize}
        \item Update Lagrange multipliers:
        \[
        \alpha^{(k+1)} = \alpha^{(k)} + \rho \cdot g(\theta^{(k+1)})
        \]
        \item Check convergence criteria. If satisfied, stop.
        \[
         g(\theta^{(k+1)}) < \epsilon \text{ or } \rho \geq \rho_{\max}
        \]
    \end{enumerate}
\end{enumerate}
\end{mdframed}

\subsection{Hilbert-space Gaussian process approximation}\label{app:hsgp}

We build on the Hilbert-space Gaussian process (HSGP) approximation introduced by Solin and Särkkä~\cite{solin2020hilbert}, which provides a computationally efficient low-rank representation of stationary Gaussian process (GP) covariance functions. 
The key idea is to project the GP onto the eigenbasis of the negative Laplacian operator defined on a bounded domain $\Omega = [-L, L]$ with Dirichlet boundary conditions. 

Let $\{\phi_j, \lambda_j\}_{j \ge 1}$ denote the eigenfunctions and eigenvalues of the Laplacian, satisfying
\begin{equation}
    -\nabla^2 \phi_j(x) = \lambda_j \, \phi_j(x),
    \qquad x \in [-L, L],
    \qquad \phi_j(\pm L) = 0.
\end{equation}
In one dimension, these eigenpairs admit closed-form expressions:
\begin{equation}
    \lambda_j = \left( \frac{\pi j}{2L} \right)^2, 
    \qquad 
    \phi_j(x) = \sqrt{\frac{1}{L}} \,
    \sin\!\left( \frac{\pi j (x + L)}{2L} \right),
    \qquad j = 1, 2, \ldots
\end{equation}
forming an orthonormal basis of $L^2([-L, L])$.

For a stationary kernel $k(x, x')$ with spectral density $S(\omega)$ (via Bochner’s theorem), the covariance function can be approximated as
\begin{equation}
    k(x, x') \approx 
    \sum_{j=1}^{m} S\!\left(\sqrt{\lambda_j}\right)
    \phi_j(x)\,\phi_j(x').
    \label{eq:hsgp_approx}
\end{equation}
Defining feature functions:
\begin{equation}
    \Phi_j(x) = \sqrt{S\!\left(\sqrt{\lambda_j}\right)}\,\phi_j(x),
\end{equation}
the kernel matrix admits a low-rank factorization
\begin{equation}
    \mathbf{K} \approx \boldsymbol{\Phi}\boldsymbol{\Phi}^{\top},
\end{equation}
where $\boldsymbol{\Phi} \in \mathbb{R}^{n \times m}$ is the feature matrix evaluated at the training inputs.

\subsubsection{Squared-exponential kernel in one dimension}

For the squared-exponential (SE) kernel
\begin{equation}
    k_{\text{SE}}(t,t') = 
    \alpha \, \exp\!\left(-\frac{(t - t')^2}{2\ell^2}\right),
\end{equation}
the corresponding spectral density is Gaussian:
\begin{equation}
    S(\omega) = \alpha \sqrt{2\pi}\, \ell \,
    \exp\!\left(-\tfrac{1}{2} \ell^2 \omega^2\right).
\end{equation}
Substituting $\omega_j = \sqrt{\lambda_j} = \pi j/(2L)$ into~\eqref{eq:hsgp_approx} yields the one-dimensional HSGP features:
\begin{equation}
    \Phi_j(t)
    = \sqrt{\alpha \sqrt{2\pi}\, \ell}\,
      \exp\!\left(-\tfrac{1}{4} \ell^2 \lambda_j \right)
      \sqrt{\tfrac{1}{L}}\,
      \sin\!\left(\tfrac{\pi j (t + L)}{2L}\right).
\end{equation}
Truncating the basis to \(m\) terms provides a controlled low-rank
approximation whose accuracy improves as \(L\) and \(m\) increase
appropriately. Under the standard regularity assumptions on the stationary
covariance and its spectral density, the approximation converges uniformly in
one dimension as \(L,m\to\infty\) with \(m/L\to\infty\).
This approximation is applied to each component of the structured covariance function (see \ref{app:inter_task})

{

\subsection{Online Woodbury updates with low-rank accumulators}
\label{app:online_batch_solver}

We consider covariance matrices of the form
\[
    K = M + \Phi\Phi^\top,
\]
where \(M\) is block-diagonal and \(\Phi\) is a low-rank feature matrix. In
\textsc{LP-StructGP}, this structure arises after applying the HSGP
approximation to the inter-subject latent pathway component. The block-diagonal
term \(M\) encodes subject-specific covariance terms, whereas the low-rank term
\(\Phi\Phi^\top\) couples subjects through shared latent pathways.

Suppose that the data is processed in mini-batches
\(\mathcal B_t\), each containing subject-level blocks
\((y_i,\Phi_i,M_i)\). For each block \(i\in\mathcal B_t\), define
\[
    A_i = M_i^{-1}y_i,
    \qquad
    B_i = M_i^{-1}\Phi_i,
\]
and the local Woodbury contributions
\[
    s_i = y_i^\top A_i,
    \qquad
    d_i = \Phi_i^\top A_i,
    \qquad
    c_i = \Phi_i^\top B_i .
\]
The batch-level contributions are
\[
    s_t = \sum_{i\in\mathcal B_t}s_i,
    \qquad
    d_t = \sum_{i\in\mathcal B_t}d_i,
    \qquad
    c_t = \sum_{i\in\mathcal B_t}c_i .
\]

Without fading, the cumulative accumulators after batch \(t\) are
\[
    S_t = S_{t-1}+s_t,
    \qquad
    D_t = D_{t-1}+d_t,
    \qquad
    C_t = C_{t-1}+c_t,
\]
with \(S_0=D_0=C_0=0\). The low-rank Woodbury correction is
\[
    E_t = (I+C_t)^{-1}D_t .
\]
By the Woodbury identity,
\[
    K_{1:t}^{-1}y_{1:t}
    =
    M_{1:t}^{-1}y_{1:t}
    -
    M_{1:t}^{-1}\Phi_{1:t}E_t,
\]
where \(K_{1:t}=M_{1:t}+\Phi_{1:t}\Phi_{1:t}^\top\). Hence the inverse-action
block for the latest mini-batch is
\[
    x_i^{(t)} = A_i - B_iE_t,
    \qquad i\in\mathcal B_t .
\]
This quantity is the current-batch block of the full inverse action associated
with the cumulative state after assimilating batches \(1,\ldots,t\).

The same accumulators also give the cumulative quadratic form and
log-determinant. Specifically,
\[
    y_{1:t}^\top K_{1:t}^{-1}y_{1:t}
    =
    S_t-D_t^\top E_t,
\]
and, by the matrix determinant lemma,
\[
    \log\det K_{1:t}
    =
    \sum_{\tau=1}^t\sum_{i\in\mathcal B_\tau}\log\det M_i
    +
    \log\det(I+C_t).
\]
Thus both the inverse action and the likelihood terms can be updated without
forming the full covariance matrix.

In the practical implementation, the accumulators may be updated with geometric
fading,
\[
    S_t = \beta S_{t-1}+s_t,
    \qquad
    D_t = \beta D_{t-1}+d_t,
    \qquad
    C_t = \beta C_{t-1}+c_t,
    \qquad 0\leq\beta<1.
\]
This implements an exponentially weighted memory of previous batches. Setting
\(\beta=0\) uses only the current batch, whereas \(\beta\to1\) approaches the
no-fading accumulator. The exact full-data finite-rank HSGP likelihood is
recovered by the no-fading reference case in which all accumulators are formed
for a fixed parameter value. With fixed \(\beta<1\), the resulting objective is
a fading-memory surrogate.

\subsection{Mini-batch conditional marginal likelihood}
\label{app:minibatch_nmll}

For \textsc{LP-StructGP}, the finite-rank HSGP approximation yields the joint
covariance
\[
    K_\theta
    =
    M_\theta+\Phi_\theta\Phi_\theta^\top,
\]
where \(M_\theta\) is block-diagonal over subject-level blocks and
\(\Phi_\theta\) contains the low-rank features coupling subjects through the
shared latent pathways. For all observations \(y\in\mathbb R^N\), the negative
log marginal likelihood is
\[
    \mathcal L(\theta)
    =
    \frac12 y^\top K_\theta^{-1}y
    +
    \frac12\log\det K_\theta
    +
    \frac{N}{2}\log(2\pi).
\]

The likelihood admits the exact chain-rule factorization
\[
    p(y_{1:T}\mid\theta)
    =
    \prod_{t=1}^T
    p(y_t\mid y_{1:t-1},\theta).
\]
Equivalently, if \(K_{1:t}\) denotes the covariance restricted to all data
observed up to batch \(t\), and
\[
    \mathcal L_{1:t}(\theta)
    =
    \frac12 y_{1:t}^\top K_{1:t}^{-1}y_{1:t}
    +
    \frac12\log\det K_{1:t}
    +
    \frac{N_{1:t}}{2}\log(2\pi),
\]
then the conditional contribution of batch \(t\) is the cumulative increment
\[
    \mathcal L_t(\theta)
    =
    \mathcal L_{1:t}(\theta)-\mathcal L_{1:t-1}(\theta).
\]
Thus
\[
\begin{aligned}
    \mathcal L_t(\theta)
    &=
    \frac12
    \left[
        y_{1:t}^\top K_{1:t}^{-1}y_{1:t}
        -
        y_{1:t-1}^\top K_{1:t-1}^{-1}y_{1:t-1}
    \right]
    \\
    &\quad+
    \frac12
    \left[
        \log\det K_{1:t}
        -
        \log\det K_{1:t-1}
    \right]
    +
    \frac{N_t}{2}\log(2\pi).
\end{aligned}
\]

Using the Woodbury accumulators from
Section~\ref{app:online_batch_solver}, define the current batch contributions
\[
    s_t
    =
    \sum_{i\in\mathcal B_t}
    y_i^\top M_i^{-1}y_i,
    \qquad
    d_t
    =
    \sum_{i\in\mathcal B_t}
    \Phi_i^\top M_i^{-1}y_i,
    \qquad
    c_t
    =
    \sum_{i\in\mathcal B_t}
    \Phi_i^\top M_i^{-1}\Phi_i .
\]
The cumulative statistics are updated as
\[
    S_t=S_{t-1}+s_t,
    \qquad
    D_t=D_{t-1}+d_t,
    \qquad
    C_t=C_{t-1}+c_t,
\]
with \(S_0=0\), \(D_0=0\), and \(C_0=0\) in the no-fading case. Let
\[
    E_t=(I+C_t)^{-1}D_t .
\]
By the Woodbury identity,
\[
    y_{1:t}^\top K_{1:t}^{-1}y_{1:t}
    =
    S_t-D_t^\top E_t .
\]
Therefore, the exact quadratic increment is
\[
\begin{aligned}
    \Delta Q_t
    &=
    \left(S_t-D_t^\top E_t\right)
    -
    \left(S_{t-1}-D_{t-1}^\top E_{t-1}\right)
    \\
    &=
    s_t
    -
    \left[
        D_t^\top(I+C_t)^{-1}D_t
        -
        D_{t-1}^\top(I+C_{t-1})^{-1}D_{t-1}
    \right].
\end{aligned}
\]
Similarly, by the matrix determinant lemma,
\[
    \log\det K_{1:t}
    =
    \sum_{\tau=1}^{t}
    \sum_{i\in\mathcal B_\tau}
    \log\det M_i
    +
    \log\det(I+C_t),
\]
so the log-determinant increment is
\[
    \Delta R_t
    =
    \sum_{i\in\mathcal B_t}\log\det M_i
    +
    \log\det(I+C_t)
    -
    \log\det(I+C_{t-1}).
\]
The mini-batch conditional NMLL contribution used for training is therefore
\[
\boxed{
    \mathcal L_t(\theta)
    =
    \frac12\Delta Q_t
    +
    \frac12\Delta R_t
    +
    \frac{N_t}{2}\log(2\pi)
}
\]
where \(N_t=\sum_{i\in\mathcal B_t}\dim(y_i)\). In the no-fading case, if
\(\theta\) is fixed during the sweep, these increments telescope:
\[
    \sum_{t=1}^T
    \mathcal L_t(\theta)
    =
    \mathcal L(\theta)
\]
for the finite-rank HSGP model.

It is useful to relate this expression to the blockwise inverse action. For
\(i\in\mathcal B_t\), the Woodbury solver gives
\[
    x_i^{(t)}
    =
    A_i-B_iE_t,
    \qquad
    A_i=M_i^{-1}y_i,
    \qquad
    B_i=M_i^{-1}\Phi_i .
\]
Hence
\[
    \sum_{i\in\mathcal B_t}
    y_i^\top x_i^{(t)}
    =
    s_t-d_t^\top E_t .
\]
Since \(D_t=D_{t-1}+d_t\), the exact quadratic increment can equivalently be
written as
\[
\begin{aligned}
    \Delta Q_t
    &=
    s_t
    -
    \left[
        D_t^\top E_t
        -
        D_{t-1}^\top E_{t-1}
    \right]
    \\
    &=
    \sum_{i\in\mathcal B_t}
    y_i^\top x_i^{(t)}
    +
    D_{t-1}^\top(E_{t-1}-E_t).
\end{aligned}
\]
The first term is the contribution of the current batch to the cumulative
quadratic form after assimilating batch \(t\). The second term is the
previous-state correction: it accounts for the fact that adding the current
batch changes the low-rank correction applied to all previously accumulated
batches. The implementation uses this corrected telescoping quadratic increment
when evaluating the training loss.

\subsection{Simulation details}\label{app:simulation_details}

We summarize here the data generation procedures and parameter settings used in the simulation study.

\subsubsection{Experiment A}

\paragraph{Graph and kernel parameters}
For each simulation, the support of $\mathbf{S}$ was sampled from a random Erd\H{o}s--R\'enyi graph with mean degree~2.
Nonzero edge weights of $\mathbf{S}$ were uniformly drawn from $[-1.5, -0.5] \cup [0.5, 1.5]$.
Log-lengthscale parameters $\log \ell_{u}$ were tied by source and uniformly sampled in $[0, 1]$.
Observation times $t$ were drawn uniformly in $[0, 10]$.
The noise variance was fixed to $\sigma^2 = 10^{-2}$ and provided as an oracle during model fitting.
The graph-recovery simulations used the SNR-preserving internally standardized StructGP parameterization.
Thus, latent task components were normalized to unit marginal variance, while the shared raw observation-noise variance was transformed using the same taskwise scales, preserving task-specific signal-to-noise ratios.

\paragraph{Latent pathway parameters}
In LP-StructGP simulations, each subject trajectory was generated as a mixture of one shared latent pathway and an individual component,
\[
\mathbf{Y}_i(t) = 0.3\, (\mathbf{G}_{i} * \mathbf{Q}_{u_i})(t) + 0.7\, (\mathbf{H} * \mathbf{w}_i)(t),
\]
where $u_i$ denotes the latent pathway assigned to subject $i$, and the coefficients $0.3$ and $0.7$ control the relative contributions of the latent and subject-specific components, respectively.
The number of
latent pathways was varied over $\{3,6,12\}$.

\paragraph{Design}
Each simulation involved $k = 10$ tasks and $25$ observations per task.
We varied the number of subjects while keeping other parameters fixed.
For each configuration, $30$ independent repetitions were performed, and the median and interquartile range of all evaluation metrics (SHD, ARI, and F1) were reported.

\paragraph{Implementation}
All simulations were implemented in \texttt{PyTorch} using the same optimization settings as described in Section 2.4: Simulation study of the manuscript.

\subsubsection{Experiment B}

Experiment B used the same graph, kernel, and noise parameter settings as Experiment A, but considered StructGP only, without latent pathways, and fixed the cohort size at 1,000 subjects.
To reproduce realistic observation patterns, uniform sampling of 25 observations was replaced by empirical sampling schedules from MIMIC-IV.
Specifically, for each simulated subject--task time series, an entire timestamp sequence was sampled with replacement from the set of observed MIMIC-IV time series; individual timestamps were not sampled independently.
The experiment was repeated 150 times.

\subsubsection{Experiment C}

Experiment C assessed recovery of the number of latent pathways under a high shared-to-individual component ratio.
We varied the true number of latent pathways from 2 to 6 and evaluated candidate models with 1 to 6 pathways.
We used the LP-StructGP settings from Experiment A with 100 subjects, with the following modifications:
\begin{itemize}
    \item the number of tasks was reduced to 3 to limit the dimensionality of each latent pathway;
    \item inter-task kernel parameters were fixed to their oracle values to reduce computational cost;
    \item the relative contributions of the shared and subject-specific components were set to \(0.7\) and \(0.3\), respectively;
    \item Log-lengthscale parameters $\log \ell_u$ were sampled uniformly from $[1,2]$; and
    \item the noise variance was fixed to \(\sigma^2 = 10^{-1}\).
\end{itemize}
Each combination of true and candidate pathway counts was evaluated over 50 independent repetitions.

\subsection{Real-life evaluation details}\label{app:real_life_details}

\paragraph{Cohort selection}
Patients were included if they met Sepsis-3 criteria: suspected infection and a Sequential Organ Failure Assessment (SOFA) score $\geq 2$.
Septic shock was defined as sepsis with persistent hypotension requiring vasopressors to maintain a mean blood pressure (MBP) $\geq 65$~mmHg and a serum lactate $> 2$~mmol$\cdot$L$^{-1}$ despite adequate fluid resuscitation.
For patients with multiple ICU stays, only the first stay was retained.

\paragraph{Data extraction and preprocessing}
Time~$t=0$ was set to ICU admission.
We extracted the first seven days of the ICU stay and excluded patients with fewer than the 20th percentile of available measurements for creatinine, norepinephrine, or MBP.
All time series were resampled into overlapping 24-hour windows for the short-term experiments.
Marginal distributions were transformed to standard normal using a quantile transform fitted on the training set.
Only strictly positive recorded vasoactive drugs doses were included as observations; periods without administration
were not encoded as zero-dose observations.

\paragraph{Experimental protocol}
The dataset was split by subject into 70\% training, 15\% validation, and 15\% test.
Each configuration was repeated five times with different initializations for unstructured baseline models.
Unless stated otherwise, all hyperparameters and optimizer settings were identical across architectures.

\paragraph{Evaluation metrics}
We report the Root Mean Square Error (RMSE) and empirical 95\% predictive coverage on test data.
Additional exploratory experiments with time-shifted features and extended covariates were also performed.

\paragraph{Extended task set}
For the large-scale evaluation, we considered 18 tasks comprising vital signs, laboratory biomarkers, and vasoactive drug administration rates:

\begin{itemize}
    \item \textbf{Hemodynamic variables:} mean blood pressure (MBP), systolic blood pressure (SBP) and heart rate (HR).
    \item \textbf{Respiratory variables:} respiratory rate (RR), oxygen saturation (SpO$_2$), partial pressure of oxygen (PO$_2$), partial pressure of carbon dioxide (pCO$_2$), blood pH, and venous oxygen saturation (sO$_2$).
    \item \textbf{Renal and metabolic biomarkers:} serum creatinine (Cr), serum lactate (Lac).
    \item \textbf{Vasoactive and therapeutic interventions:} norepinephrine (NE), epinephrine (Epi), vasopressin (Vaso), phenylephrine (PE), dobutamine (Dobut), and sodium chloride 0.9\% (NaCl).
    \item \textbf{others}: body temperature (Temp)
\end{itemize}

All variables were preprocessed and standardized following the same procedure as described above.

\section{Supplementary results}
\subsection{Missingness}\label{app:missingness}

\begin{figure}[hbt!]
  \centering
  \includegraphics[width=0.9\textwidth]{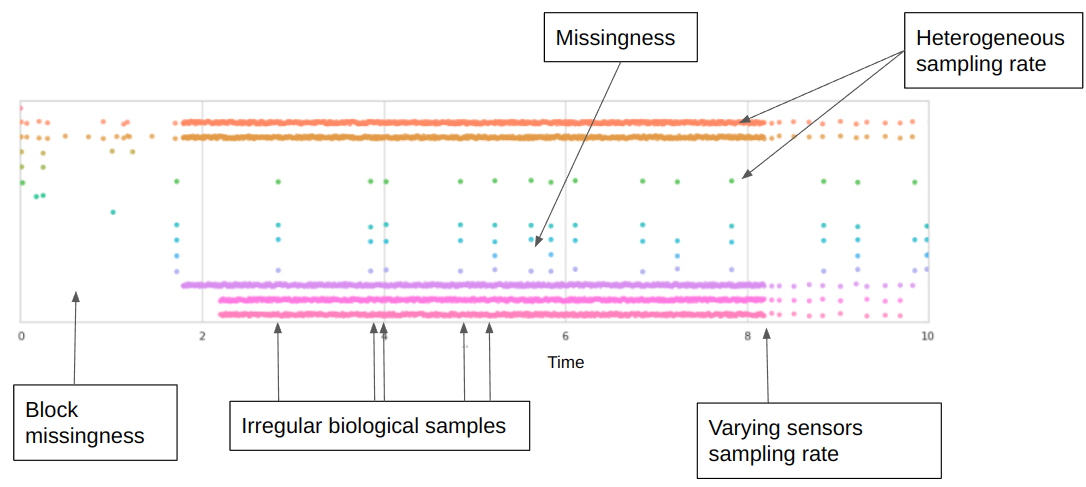}
  \caption{A patient's stay observation pattern in Electronic Health Records.\\
  \footnotesize
  \\
  Scatter plot of a patient's stay, with one point per observation.
  Each line/color represents a different variable (y-axis) as a function of time in days (x-axis). }
  \label{fig:messy}
\end{figure}

\subsection{Simulation}\label{app:simulation}

\begin{figure}[hbt!]
  \centering
  \includegraphics[width=1.\textwidth]{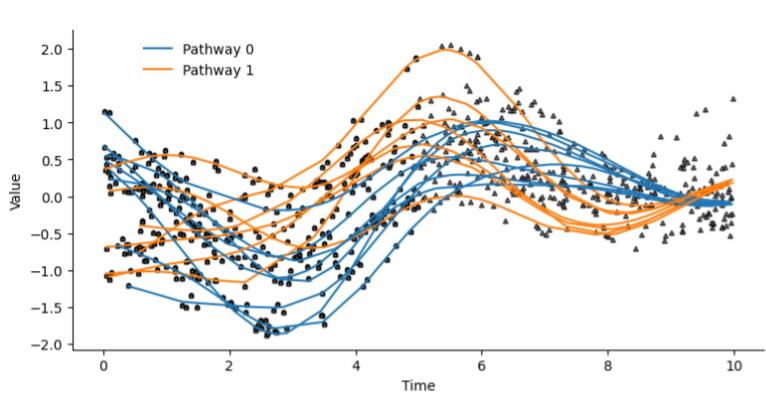}
\caption[Forecasting after uncovering latent pathways]{Forecasting after uncovering latent pathways.\\
\small
Toy model illustrating the interpolation between group and individual forecasts after conditioning on observations from $t=0$ to $t=5$, once latent pathways have been uncovered.
}
  \label{fig:toy_tsm}
\end{figure}

\subsection{PhysioNet Challenge}\label{app:physionet}

\begin{table}[hbt!]
    \centering
\resizebox{\textwidth}{!}{
\begin{tabular}{lllll}
\toprule
 & \multicolumn{2}{r}{\textbf{MAE $\times 10^{-2}$ [95\% CI]}} & \multicolumn{2}{r}{\textbf{MSE $\times 10^{-3}$ [95\% CI]}} \\
 & GraFITi & StructGP (best) & GraFITi & StructGP (best) \\
\midrule
GCS & 9.48 [9.30, 9.65] & 13.71 [13.44, 14.01] & 24.67 [23.75, 25.65] & 59.73 [57.65, 61.86] \\
FiO2 & 10.55 [10.25, 10.86] & 12.61 [12.27, 12.96] & 28.88 [27.30, 30.43] & 42.17 [40.19, 44.21] \\
Cholesterol & 8.62 [5.75, 13.00] & 9.82 [7.18, 13.43] & 16.41 [4.70, 40.74] & 16.62 [6.82, 34.11] \\
TroponinT & 2.19 [1.59, 2.95] & 3.00 [2.20, 3.82] & 4.07 [1.10, 8.45] & 6.72 [2.92, 11.54] \\
Lactate & 3.49 [3.24, 3.74] & 4.07 [3.72, 4.45] & 3.79 [3.06, 4.53] & 6.24 [4.75, 7.89] \\
TroponinI & 11.92 [9.10, 15.68] & 12.39 [8.27, 16.25] & 31.36 [16.17, 55.13] & 42.85 [18.77, 69.23] \\
PaO2 & 7.16 [6.94, 7.39] & 7.46 [7.23, 7.70] & 11.88 [10.83, 12.96] & 13.48 [12.33, 14.76] \\
SysABP & 5.22 [5.17, 5.26] & 5.51 [5.46, 5.56] & 4.72 [4.61, 4.82] & 5.40 [5.28, 5.51] \\
HCT & 4.26 [4.16, 4.37] & 4.52 [4.41, 4.63] & 3.14 [2.96, 3.31] & 3.55 [3.35, 3.75] \\
Bilirubin & 1.79 [1.54, 2.07] & 2.02 [1.75, 2.33] & 1.38 [0.94, 1.88] & 1.84 [1.33, 2.45] \\
NISysABP & 4.37 [4.33, 4.42] & 4.56 [4.51, 4.60] & 3.39 [3.30, 3.50] & 3.74 [3.63, 3.84] \\
BUN & 2.44 [2.35, 2.53] & 2.58 [2.47, 2.67] & 1.35 [1.22, 1.51] & 1.54 [1.37, 1.73] \\
NIMAP & 4.22 [4.17, 4.26] & 4.33 [4.29, 4.37] & 3.03 [2.95, 3.11] & 3.24 [3.16, 3.31] \\
Albumin & 8.71 [8.03, 9.40] & 8.82 [8.14, 9.49] & 13.46 [11.30, 15.67] & 13.78 [11.75, 15.94] \\
HR & 3.18 [3.15, 3.20] & 3.27 [3.25, 3.30] & 1.91 [1.88, 1.95] & 2.07 [2.03, 2.11] \\
HCO3 & 4.29 [4.18, 4.42] & 4.38 [4.24, 4.49] & 3.29 [3.04, 3.56] & 3.50 [3.22, 3.81] \\
DiasABP & 2.82 [2.79, 2.84] & 2.87 [2.85, 2.90] & 1.47 [1.42, 1.53] & 1.63 [1.57, 1.69] \\
Platelets & 1.40 [1.35, 1.45] & 1.44 [1.39, 1.50] & 0.43 [0.37, 0.52] & 0.43 [0.38, 0.51] \\
\bottomrule
\end{tabular}
}
\caption{Underperforming tasks compared with GraFITi (SOTA) on the PhysioNet challenge
\\\footnotesize
Mean absolute error (MAE) and Mean squared error (MSE) with bootstrap 95\% confidence intervals. Metrics are reported for up to a 1-day forecasting horizon for 18 tasks, after the first 24 hours of the ICU stay.
}
\label{tab:underperforming_tasks}
\end{table}

\begin{table}[hbt!]
    \centering
\resizebox{\textwidth}{!}{
\begin{tabular}{lllll}
\toprule
 & \multicolumn{2}{r}{\textbf{MAE $\times 10^{-2}$ [95\% CI]}} & \multicolumn{2}{r}{\textbf{MSE $\times 10^{-3}$ [95\% CI]}} \\
 & GraFITi & StructGP (best) & GraFITi & StructGP (best) \\
\midrule
PaCO2 & 4.77 [4.65, 4.89] & 4.78 [4.65, 4.91] & 4.29 [4.02, 4.58] & 4.50 [4.16, 4.87] \\
MBP & 3.43 [3.39, 3.46] & 3.42 [3.39, 3.46] & 2.21 [2.13, 2.31] & 2.38 [2.28, 2.48] \\
Na & 3.00 [2.91, 3.09] & 2.98 [2.89, 3.07] & 1.64 [1.52, 1.77] & 1.64 [1.52, 1.79] \\
NIDiasABP & 4.31 [4.26, 4.35] & 4.28 [4.24, 4.32] & 3.26 [3.18, 3.34] & 3.31 [3.22, 3.39] \\
RespRate & 3.27 [3.22, 3.32] & 3.23 [3.19, 3.28] & 1.96 [1.87, 2.07] & 1.98 [1.88, 2.10] \\
Creatinine & 1.23 [1.16, 1.29] & 1.19 [1.12, 1.25] & 0.52 [0.39, 0.73] & 0.52 [0.44, 0.61] \\
K & 1.85 [1.79, 1.92] & 1.78 [1.72, 1.83] & 0.70 [0.61, 0.81] & 0.60 [0.55, 0.65] \\
ALP & 0.72 [0.58, 0.93] & 0.60 [0.46, 0.80] & 0.50 [0.08, 1.28] & 0.49 [0.06, 1.23] \\
Glucose & 1.96 [1.88, 2.03] & 1.80 [1.73, 1.87] & 0.87 [0.73, 1.00] & 0.81 [0.66, 0.96] \\
ALT & 1.43 [1.17, 1.71] & 1.27 [0.98, 1.65] & 1.59 [0.90, 2.45] & 1.85 [0.95, 2.96] \\
Temp & 1.03 [1.01, 1.06] & 0.86 [0.84, 0.88] & 0.41 [0.27, 0.57] & 0.35 [0.21, 0.52] \\
SaO2 & 1.84 [1.74, 1.96] & 1.61 [1.50, 1.73] & 0.87 [0.61, 1.24] & 0.99 [0.69, 1.37] \\
AST & 1.31 [1.09, 1.57] & 1.09 [0.83, 1.37] & 1.34 [0.71, 2.26] & 1.52 [0.77, 2.41] \\
Mg & 0.83 [0.80, 0.85] & 0.60 [0.58, 0.63] & 0.12 [0.10, 0.15] & 0.08 [0.06, 0.11] \\
Urine & 0.84 [0.83, 0.85] & 0.59 [0.58, 0.60] & 0.13 [0.12, 0.14] & 0.13 [0.12, 0.14] \\
WBC & 0.36 [0.36, 0.37] & 0.04 [0.04, 0.04] & 0.02 [0.02, 0.02] & 0.00 [0.00, 0.00] \\
Weight & 0.67 [0.67, 0.68] & 0.32 [0.31, 0.33] & 0.14 [0.12, 0.15] & 0.11 [0.09, 0.12] \\
pH & 0.37 [0.36, 0.39] & 0.01 [0.01, 0.01] & 0.03 [0.03, 0.03] & 0.00 [0.00, 0.00] \\
MechVent & 0.47 [0.46, 0.47] & 0.00 [0.00, 0.00] & 0.03 [0.03, 0.04] & 0.00 [0.00, 0.00] \\
\bottomrule
\end{tabular}
}
\caption{Outperforming tasks compared with GraFITi (SOTA) on the PhysioNet challenge
\\\footnotesize
Mean absolute error (MAE) and Mean squared error (MSE) with bootstrap 95\% confidence intervals. Metrics are reported for up to a 1-day forecasting horizon for 19 tasks, after the first 24 hours of the ICU stay.
}
\label{tab:outperforming_tasks}
\end{table}

\begin{table}[hbt!]
    \centering
\resizebox{\textwidth}{!}{
\begin{tabular}{llll}
\toprule
& & \textbf{MAE $\times 10^{-2}$ [95\% CI]} &  \textbf{MSE $\times 10^{-3}$ [95\% CI]} \\
\midrule
\multicolumn{2}{l}{\textbf{High Model Certainty}} \\
 & GraFITi & 2.53 [2.44, 2.64]  & 2.03 [1.81, 2.30]  \\
 & StructGP (best) & 2.46 [2.34, 2.60]  & 2.20 [1.90, 2.58]  \\
\multicolumn{2}{l}{\textbf{Moderate Model Certainty}} \\
 & GraFITi & 2.81 [2.62, 3.03]  & 2.97 [2.06, 4.01]  \\
 & StructGP (best) & 2.96 [2.73, 3.20]  & 3.79 [2.81, 5.00]  \\
\multicolumn{2}{l}{\textbf{Low Model Certainty}} \\
 & GraFITi & 2.80 [2.71, 2.90]  & 2.27 [2.08, 2.47]  \\
 & StructGP (best) & 3.05 [2.85, 3.27]  & 3.59 [2.78, 4.62]  \\
% \midrule
\bottomrule
\end{tabular}
}
\caption{Performance Comparison across model confidence levels on the PhysioNet challenge
\\\footnotesize
Mean absolute error (MAE) and Mean squared error (MSE) with bootstrap 95\% confidence intervals. Metrics are reported for up to a 1-day forecasting horizon for 41 tasks, after the first 24 hours of the ICU stay.
Metrics are segmented into terciles based on the StructGP (best) model posterior predictive variance.
}
\label{tab:postvar}
\end{table}

% Required packages:
% \usepackage{booktabs}
% \usepackage{tabularx}
% \usepackage{array}
\begin{table}[hbt!]
\centering

\footnotesize
\setlength{\tabcolsep}{3.2pt}
\renewcommand{\arraystretch}{1.15}
\begin{tabularx}{\textwidth}{
    >{\raggedright\arraybackslash}p{5cm}
    >{\centering\arraybackslash}p{1.cm}
    >{\centering\arraybackslash}p{1.cm}
    >{\centering\arraybackslash}p{1.cm}
    >{\centering\arraybackslash}p{1.cm}
    >{\centering\arraybackslash}p{1.cm}
    >{\centering\arraybackslash}p{1.3cm}
}
\toprule
& &
\multicolumn{4}{c}{Grouped by latent pathway}
& \\
\cmidrule(lr){3-6}
\textbf{Characteristic}
& \textbf{Overall}
& \textbf{Unk.}
& \textbf{lp0}
& \textbf{lp1}
& \textbf{lp2}
& \textbf{P-value} \\
\midrule

$n$ & 705 & 309 & 107 & 130 & 159 & \\

Age, years, median [Q1, Q3]
& 67 [56,77] & 67 [56,77] & 66 [56,76] & 63 [53,74]
& 72 [61,80] & 0.002 \\

Male, n (\%)
& 404 (57.3) & 184 (59.5) & 41 (38.3) & 89 (68.5)
& 90 (56.6) & $<0.001$ \\

SOFA score, median [Q1, Q3]
& 5.0 [4.0,8.0] & 5.0 [4.0,8.0] & 5.0 [4.0,6.0]
& 5.0 [4.0,8.0] & 5.0 [4.0,8.0] & 0.007 \\

History of myocardial infarction, n (\%)
& 111 (15.7) & 51 (16.5) & 13 (12.1) & 19 (14.6)
& 28 (17.6) & 0.633 \\

Baseline serum creatinine, mg/dL, median [Q1, Q3]
& 0.8 [0.6,1.0] & 0.8 [0.6,1.0] & 0.5 [0.4,0.6]
& 0.9 [0.8,1.1] & 1.0 [0.8,1.1] & $<0.001$ \\

Charlson Comorbidity Index, median [Q1, Q3]
& 5.0 [3.0,7.0] & 5.0 [3.0,7.0] & 5.0 [2.0,6.0]
& 5.0 [3.0,6.0] & 6.0 [4.0,7.5] & $<0.001$ \\

In-hospital mortality, n (\%)
& 235 (33.3) & 92 (29.8) & 32 (29.9) & 36 (27.7)
& 75 (47.2) & $<0.001$ \\

Congestive heart failure, n (\%)
& 185 (26.2) & 81 (26.2) & 17 (15.9) & 40 (30.8)
& 47 (29.6) & 0.042 \\

Peripheral vascular disease, n (\%)
& 81 (11.5) & 35 (11.3) & 9 (8.4) & 17 (13.1)
& 20 (12.6) & 0.679 \\

Cerebrovascular disease, n (\%)
& 54 (7.7) & 21 (6.8) & 8 (7.5) & 10 (7.7)
& 15 (9.4) & 0.792 \\

Chronic pulmonary disease, n (\%)
& 161 (22.8) & 71 (23.0) & 23 (21.5) & 26 (20.0)
& 41 (25.8) & 0.684 \\

Mild liver disease, n (\%)
& 110 (15.6) & 41 (13.3) & 10 (9.3) & 25 (19.2)
& 34 (21.4) & 0.020 \\

Moderate or severe liver disease, n (\%)
& 43 (6.1) & 17 (5.5) & 4 (3.7) & 5 (3.8)
& 17 (10.7) & 0.041 \\

Diabetes without end-organ damage, n (\%)
& 151 (21.4) & 60 (19.4) & 21 (19.6) & 30 (23.1)
& 40 (25.2) & 0.480 \\

Diabetes with end-organ damage, n (\%)
& 48 (6.8) & 27 (8.7) & 0 (0.0) & 7 (5.4)
& 14 (8.8) & 0.011 \\

Chronic kidney disease, n (\%)
& 92 (13.0) & 37 (12.0) & 1 (0.9) & 23 (17.7)
& 31 (19.5) & $<0.001$ \\

Malignancy, n (\%)
& 67 (9.5) & 33 (10.7) & 11 (10.3) & 9 (6.9)
& 14 (8.8) & 0.644 \\

Metastatic solid tumor, n (\%)
& 27 (3.8) & 9 (2.9) & 8 (7.5) & 3 (2.3)
& 7 (4.4) & 0.137 \\

\bottomrule
\end{tabularx}
\vspace{2pt}

\caption{\textbf{Baseline characteristics according to latent pathway assignment.}
\\
Baseline demographic and clinical characteristics stratified
by latent pathway  on the training set.
Patients with a maximum posterior pathway
membership probability below 0.5 were classified as \emph{Unknown} (Unk.).
Continuous variables are reported as median [Q1, Q3] and compared using the
Kruskal--Wallis test. Categorical variables are reported as n (\%) and compared
using the chi-squared test. P-values correspond to comparisons across all
pathway groups (Unknown, lp0, lp1, and lp2). SOFA, Sequential Organ Failure
Assessment.
}
\label{tab:pathwaydesc}
\end{table}

\clearpage

\section{Analysis}
\subsection{Identifiability of the StructGP ordered-dependence graph from the marginal likelihood}
\label{app:identifiability_structgp}

We establish here a population identifiability result for the ordered
inter-task dependency graph learned by \textsc{StructGP}.  The argument works
directly with the Gaussian process-convolution factor and does not require an
additional structural-equation representation of the spectral density.

\paragraph{Raw StructGP model and graph.}
Let
\[
    Y(t)=(Y_1(t),\ldots,Y_k(t))^\top
\]
be a zero-mean stationary Gaussian process generated by
\[
    Y(t)=(H*w)(t),
\]
where the components of \(w(t)\) are mutually independent Gaussian white
noises with unit spectral density.  The entries of the matrix-valued filter are
\[
    H_{vu,\theta}(t)
    =
    S_{vu,\theta}
    \exp\!\left(-\frac{t^2}{\ell_{vu,\theta}}\right),
    \qquad
    \ell_{vu,\theta}>0.
\]
Following the convention introduced in
Appendix~\ref{app:inter_task}, \(S\) denotes the full amplitude matrix
including the fixed identity self-connections. Hence
\[
    S_{vv,\theta}=1,
    \qquad v=1,\ldots,k.
\]
Only the off-diagonal entries are learned as graph coefficients.  Their
support is assumed to be acyclic, or equivalently lower triangular after some
permutation of the tasks.  The corresponding ordered-dependence graph is
\[
    \mathcal G_H(\theta)
    =
    \big(\{1,\ldots,k\},\mathcal E_H(\theta)\big),
\]
where
\[
    u\to v\in\mathcal E_H(\theta)
    \quad\Longleftrightarrow\quad
    H_{vu,\theta}\not\equiv0
    \quad\Longleftrightarrow\quad
    S_{vu,\theta}\neq0,
    \qquad u\neq v.
\]
The arrow describes the ordering encoded by the triangular convolution factor;
it is not given a causal interpretation here.

With the Fourier-transform convention
\(
\widehat f(\omega)=\int_{\mathbb R}f(t)e^{-i\omega t}\,dt
\), define
\[
    h_{vu,\theta}(\omega)
    :=
    \widehat H_{vu,\theta}(\omega)
    =
    S_{vu,\theta}\sqrt{\pi\ell_{vu,\theta}}
    \exp\!\left(-\frac{\ell_{vu,\theta}\omega^2}{4}\right).
\]
Writing \(h_{u,\theta}(\omega)\) for column \(u\) of
\(\widehat H_\theta(\omega)\), the latent spectral density is
\[
    F_\theta(\omega)
    =
    \widehat H_\theta(\omega)
    \widehat H_\theta(\omega)^*
    =
    \sum_{u=1}^k
    h_{u,\theta}(\omega)h_{u,\theta}(\omega)^*.
\]
Because the Gaussian filters are real and even, these quantities are real, but
we retain the Hermitian-transpose notation.  Entrywise,
\[
\begin{aligned}
    F_{\theta,vv'}(\omega)
    &={}
    \pi\sum_{u=1}^k
    S_{vu,\theta}S_{v'u,\theta}
    \sqrt{\ell_{vu,\theta}\ell_{v'u,\theta}}
    \\
    &\qquad\times
    \exp\!\left(
    -\frac{(\ell_{vu,\theta}+\ell_{v'u,\theta})\omega^2}{4}
\right),
\end{aligned}
\]
and, in particular,
\begin{equation}
    F_{\theta,vv}(\omega)
    =
    \pi\sum_{u=1}^k
    S_{vu,\theta}^2\ell_{vu,\theta}
    \exp\!\left(-\frac{\ell_{vu,\theta}\omega^2}{2}\right).
    \label{eq:structgp_diagonal_spectrum}
\end{equation}

The following elementary fact will be used repeatedly.  If
\(a_1,\ldots,a_m\) are distinct, then the functions
\(x\mapsto e^{-a_jx}\) are linearly independent on every nonempty interval.
Indeed, differentiating a vanishing linear combination at an interior point
\(x_0\) through
order \(m-1\) gives a Vandermonde system.  Consequently, a finite positive
mixture of such exponentials is a single exponential if and only if all of its
active rates coincide.

\paragraph{Direct recovery of the convolution factor.}
The process-convolution construction guarantees existence of the displayed
factorization.  The next proposition establishes its uniqueness within the
actual StructGP parameter class.

\begin{proposition}[Uniqueness of the anchored Gaussian StructGP factor]
\label{prop:structgp_factor_uniqueness}
Suppose that a spectral density \(F\) admits a factorization
\(F=\widehat H\widehat H^*\), where the entries of \(H\) are centered Gaussian
filters, the diagonal amplitudes satisfy \(S_{vv}=1\), and the off-diagonal
support is acyclic.  Then this factor is unique within that class.  Therefore
\(F\) identifies \(\mathcal G_H=\operatorname{supp}_{\rm off}H\), every
nonzero amplitude, and the length-scale of every active filter.  Length-scales
attached to zero amplitudes are not identified, because they do not affect
\(H\) or \(F\).
\end{proposition}

\begin{proof}

Let \(\theta\) denote one admissible parameterization of \(F\).

Let \(A\subseteq V:=\{1,\ldots,k\}\) be a set of source columns that have
already been recovered and removed, and put \(R=V\setminus A\).  The
recursively deflated spectrum on the remaining vertices is
\begin{equation}
\begin{aligned}
    G^{(R)}(\omega)
    &:={}
    F_{\theta,RR}(\omega)
    -
    \sum_{u\in A}
    h_{R u,\theta}(\omega)h_{R u,\theta}(\omega)^*
    \\
    &=
    \widehat H_{\theta,RR}(\omega)
    \widehat H_{\theta,RR}(\omega)^*.
    \label{eq:structgp_residual_spectrum}
\end{aligned}
\end{equation}
At the first stage, \(A=\varnothing\), \(R=V\), and
\(G^{(V)}=F_\theta\).  For \(r\in R\), the residual diagonal entry is
\begin{equation}
\begin{aligned}
    G^{(R)}_{rr}(\omega)
    ={}&
    \pi\ell_{rr,\theta}
    \exp\!\left(-\frac{\ell_{rr,\theta}\omega^2}{2}\right)
    \\
    &+
    \pi
    \sum_{u\in\operatorname{pa}_{\mathcal G_H(\theta)}(r)\cap R}
    S_{ru,\theta}^2\ell_{ru,\theta}
    \exp\!\left(-\frac{\ell_{ru,\theta}\omega^2}{2}\right).
    \label{eq:structgp_residual_diagonal}
\end{aligned}
\end{equation}
We claim that \(r\) is a root of the induced graph on \(R\) if and only if
there exists an \(\ell>0\) such that
\begin{equation}
    G^{(R)}_{rr}(\omega)
    =
    \pi\ell\exp\!\left(-\frac{\ell\omega^2}{2}\right)
    \quad\text{for all }\omega.
    \label{eq:structgp_root_signature}
\end{equation}
The forward implication follows immediately from
\eqref{eq:structgp_residual_diagonal}.  Conversely, linear independence of
the exponentials implies that every active length-scale on the right-hand
side of \eqref{eq:structgp_residual_diagonal} must equal \(\ell\).  Equality
of the coefficients then gives
\[
    1+
    \sum_{u\in\operatorname{pa}_{\mathcal G_H(\theta)}(r)\cap R}
    S_{ru,\theta}^2
    =1.
\]
Thus \(r\) has no parent in \(R\), proving the claim.  Notice that this
criterion remains valid when several active filters have the same
length-scale: positivity prevents cancellation, while the fixed unit
self-filter fixes the coefficient of the diagonal component.

For every root \(r\) identified by \eqref{eq:structgp_root_signature}, the
positive-diagonal convention gives
\[
    h_{rr,\theta}(\omega)
    =
    \sqrt{G^{(R)}_{rr}(\omega)}.
\]
Moreover, row \(r\) contains no other unremoved source, and hence, for every
\(v\in R\),
\[
    G^{(R)}_{vr}(\omega)
    =
    h_{vr,\theta}(\omega)\overline{h_{rr,\theta}(\omega)}.
\]
Therefore the remaining part of column \(r\) is identified through
\begin{equation}
    h_{vr,\theta}(\omega)
    =
    \frac{G^{(R)}_{vr}(\omega)}{\overline{h_{rr,\theta}(\omega)}},
    \qquad v\in R.
    \label{eq:structgp_recover_root_column}
\end{equation}
For \(R'=R\setminus\{r\}\), the next residual is
\begin{equation}
\begin{aligned}
    G^{(R')}(\omega)
    &:={}
    G^{(R)}_{R'R'}(\omega)
    -
    h_{R'r,\theta}(\omega)h_{R'r,\theta}(\omega)^*
    \\
    &=
    \widehat H_{\theta,R'R'}(\omega)
    \widehat H_{\theta,R'R'}(\omega)^*.
    \label{eq:structgp_residual_update}
\end{aligned}
\end{equation}
There are no missing entries of column \(r\) on vertices removed earlier:
when such a vertex was removed, it was a root while \(r\) was still present,
so it could not have \(r\) as a parent.  Every nonempty finite DAG has a
root; recursion therefore terminates after at most \(k\) stages.  Applying
the same observable root test and update to any competing admissible factor
recovers the same columns at every stage, proving uniqueness of
\(\widehat H_\theta\) within the anchored acyclic Gaussian class.  If several
roots are present, there are no edges between them, and their rank-one
removals commute.

For every nonzero recovered entry, its Gaussian form uniquely determines its
length-scale and amplitude.  For any \(\omega\neq0\),
\begin{equation}
    \ell_{vu,\theta}
    =
    -\frac{4}{\omega^2}
    \log\!\left(
        \frac{|h_{vu,\theta}(\omega)|}{|h_{vu,\theta}(0)|}
    \right),
    \qquad
    S_{vu,\theta}
    =
    \frac{h_{vu,\theta}(0)}{\sqrt{\pi\ell_{vu,\theta}}}.
    \label{eq:structgp_recover_filter_parameters}
\end{equation}
Thus the spectral density identifies the support of \(H\) and all amplitudes
and length-scales attached to active filters.  If \(S_{vu,\theta}=0\), the
corresponding \(\ell_{vu,\theta}\) has no effect on the filter or covariance
and is necessarily nonidentifiable.  This inactive length-scale
nonidentifiability is immaterial for graph recovery.
\end{proof}

\paragraph{From the population marginal likelihood to the spectrum.}
For subject \(i\), let
\[
    \mathbf y_i \mid \mathbf X_i,\theta
    \sim
    \mathcal N\!\bigl(0,C_\theta(\mathbf X_i)\bigr),
\]
where \(C_\theta(\mathbf X_i)\) is induced by the StructGP covariance
\(k_\theta\), together with an observation-noise term that is absent,
known, or separately identifiable. Assume that the data-generating
parameter \(\theta_0\) belongs to the anchored acyclic StructGP class,
that the expected Gaussian log-likelihood is well defined, and that the
distribution of the observation designs is sufficiently rich that
\[
    C_\theta(\mathbf X)=C_{\theta'}(\mathbf X)
    \quad\text{almost surely}
    \qquad\Longrightarrow\qquad
    k_\theta=k_{\theta'}.
\]
For the Gaussian filters considered here, this condition is satisfied,
for example, when the observed lags for every pair of tasks contain a
nonempty interval: the covariance functions are analytic, so equality
on such an interval implies equality everywhere.

Let
\[
    M(\theta)
    :=
    \mathbb E_{\theta_0}
    \left[
        \log p_\theta(\mathbf y\mid\mathbf X)
    \right]
\]
denote the population marginal log-likelihood.

\begin{proposition}[Graph identifiability from the population marginal likelihood]
\label{prop:structgp_identifiability_marginal_likelihood}
Under the preceding assumptions, every maximizer of \(M\) induces the
same ordered-dependence graph as \(\theta_0\).
\end{proposition}

\begin{proof}
For every candidate parameter \(\theta\),
\[
    M(\theta_0)-M(\theta)
    =
    \mathbb E_{\mathbf X}
    \left[
        \operatorname{KL}
        \left(
            \mathcal N(0,C_{\theta_0}(\mathbf X))
            \,\middle\|\,
            \mathcal N(0,C_\theta(\mathbf X))
        \right)
    \right]
    \geq 0.
\]
Equality at a population maximizer implies
\(C_\theta(\mathbf X)=C_{\theta_0}(\mathbf X)\) almost surely. By the
design-richness and noise-identifiability assumptions, this gives
\(k_\theta=k_{\theta_0}\), and therefore equality of their spectral
density matrices. Proposition~\ref{prop:structgp_factor_uniqueness}
then implies equality of the anchored spectral factors and hence of
their ordered-dependence graphs.
\end{proof}

\begin{corollary}[Scale invariance of raw ordered-graph identifiability]
\label{cor:raw_structgp_scale_invariance}
Let \(\theta\) and \(\theta'\) be two admissible raw StructGP parameters with
fixed unit self-filter amplitudes.  If, for some \(a>0\),
\[
    k_{\theta'}(\tau)
    =
    a^2 k_\theta(\tau)
    \qquad\text{for all }\tau,
\]
then \(a=1\) and
\[
    \mathcal G_H(\theta')
    =
    \mathcal G_H(\theta).
\]
In particular, allowing an unknown global positive scale does not introduce a
graph ambiguity within the admissible raw StructGP family.
\end{corollary}

\begin{proof}
Fourier transformation gives \(F_{\theta'}=a^2F_\theta\).  Choose a root \(r\)
of \(\mathcal G_H(\theta)\).  Its diagonal spectrum under \(\theta\) is
\[
    F_{\theta,rr}(\omega)
    =
    \pi\ell_{rr,\theta}
    e^{-\ell_{rr,\theta}\omega^2/2}.
\]
Consequently, \(F_{\theta',rr}\) is a single Gaussian with coefficient
multiplied by \(a^2\).  In any admissible raw StructGP factorization, a
diagonal spectrum can be a single Gaussian only if all of its active filters
have the same length-scale.  Linear independence therefore forces every active
length-scale in row \(r\) under \(\theta'\), including
\(\ell_{rr,\theta'}\), to equal \(\ell_{rr,\theta}\).  Equality of the
coefficients gives
\[
    a^2\pi\ell_{rr,\theta}
    =
    \pi\ell_{rr,\theta}
    \left[
        1+
        \sum_{u\in\operatorname{pa}_{\mathcal G_H(\theta')}(r)}
        S_{ru,\theta'}^2
    \right]
    \geq
    \pi\ell_{rr,\theta}.
\]
It follows that \(a^2\geq1\).  Reversing the roles of \(\theta\) and
\(\theta'\), and applying the same argument to a root of
\(\mathcal G_H(\theta')\), gives \(a^{-2}\geq1\).  Hence \(a=1\).
Proposition~\ref{prop:structgp_factor_uniqueness} then gives equality of the
factors, and therefore of the graphs.
\end{proof}

\paragraph{Remark on finite-sample graph recovery.}
Proposition~\ref{prop:structgp_identifiability_marginal_likelihood}
is a population identifiability result, rather than a finite-sample
graph-selection theorem. Exact recovery from an empirical marginal
likelihood would additionally require suitable convergence of the
parameter estimator, a separation condition on nonzero edge
amplitudes, and successful optimization of the nonconvex objective.
We do not establish these additional conditions here.

\subsection{Identifiability under SNR-preserving internal standardization}
\label{app:snr_preserving_identifiability}

We now show that the internal standardization used in \textsc{StructGP} does
not destroy identifiability of the ordered-dependence graph
\[
    \mathcal G_H
    =
    \operatorname{supp}_{\rm off}H
    =
    \operatorname{supp}_{\rm off}S,
\]
provided that the standardization is applied to the full raw observation model
rather than only to the latent process. The key point is that
SNR-preserving standardization keeps the information carried by the shared raw
observation-noise scale. This information rules out taskwise rescalings that
could otherwise make different ordered-dependence graphs observationally
equivalent.

Throughout this subsection, the raw StructGP ordered-dependence graph is
defined directly by the support of the off-diagonal convolution filters:
\[
    \mathcal G_H(\theta)
    :=
    \operatorname{supp}_{\rm off}H_\theta
    =
    \operatorname{supp}_{\rm off}S_\theta.
\]
Equivalently,
\[
    u\to v \text{ in } \mathcal G_H(\theta)
    \quad\Longleftrightarrow\quad
    H_{vu,\theta}\not\equiv0
    \quad\Longleftrightarrow\quad
    S_{vu,\theta}\neq0,
    \qquad u\neq v.
\]
By Proposition~\ref{prop:structgp_factor_uniqueness}, this graph is identified
by the raw StructGP spectral density.

\paragraph{Raw latent and observation model.}
Let
\[
    \mathbf X^\theta(t)
    =
    (X^\theta_1(t),\ldots,X^\theta_k(t))^\top
\]
be the raw latent StructGP process with covariance function
\[
    \Gamma_X^\theta(\tau)
    =
    \operatorname{Cov}
    \{
        \mathbf X^\theta(t),
        \mathbf X^\theta(t+\tau)
    \}.
\]
The corresponding raw observation model is
\[
    \mathbf Y^\theta(t)
    =
    \mathbf X^\theta(t)
    +
    \boldsymbol\varepsilon(t),
    \qquad
    \boldsymbol\varepsilon(t)
    \sim
    \mathcal N(0,\sigma^2I_k),
\]
where the raw observation-noise variance is shared across tasks and satisfies
\[
    \sigma^2>0 .
\]
Thus
\[
    \Gamma_Y^\theta(\tau)
    =
    \Gamma_X^\theta(\tau)
    +
    \sigma^2I_k\mathbf 1_{\{\tau=0\}} .
\]
The assumption that the raw observation noise is task-independent is essential.
It is not a claim that the standardized model has task-independent noise.
After internal standardization, the same shared raw noise becomes
task-specific in standardized units.

\paragraph{Internal latent scales.}
For each task \(v\), define the raw latent marginal variance
\[
    q_v^\theta
    :=
    \operatorname{Var}\{X_v^\theta(t)\}
    =
    \Gamma_{X,vv}^\theta(0),
\]
and assume
\[
    q_v^\theta>0
    \qquad
    \text{for all }v=1,\ldots,k .
\]
Let
\[
    D_\theta
    =
    \operatorname{diag}(s_1^\theta,\ldots,s_k^\theta),
    \qquad
    s_v^\theta
    =
    \sqrt{q_v^\theta}.
\]
For Gaussian convolution filters of the form
\[
    H_{vu,\theta}(t)
    =
    S_{vu,\theta}
    \exp\!\left(-\frac{t^2}{\ell_{vu,\theta}}\right),
\]
one has
\[
    q_v^\theta
    =
    \sum_{u=1}^k
    S_{vu,\theta}^2
    \sqrt{\frac{\pi\ell_{vu,\theta}}{2}} .
\]

The internally standardized latent process is
\[
    \widetilde{\mathbf X}^{\theta}(t)
    =
    D_\theta^{-1}\mathbf X^\theta(t),
\]
and therefore
\[
    \operatorname{Var}
    \{
        \widetilde X_v^\theta(t)
    \}
    =
    1
    \qquad
    \text{for all }v.
\]
At the level of convolution filters,
\[
    \widetilde H_{vu,\theta}(t)
    =
    \frac{1}{s_v^\theta}H_{vu,\theta}(t).
\]
Since \(s_v^\theta>0\), this row rescaling preserves zeros:
\[
    \widetilde H_{vu,\theta}\equiv0
    \quad\Longleftrightarrow\quad
    H_{vu,\theta}\equiv0.
\]

Thus internal standardization does not itself change the
ordered-dependence graph:
\[
    \operatorname{supp}_{\rm off}\widetilde H_\theta
    =
    \operatorname{supp}_{\rm off}H_\theta
    =
    \operatorname{supp}_{\rm off}S_\theta.
\]

\paragraph{SNR-preserving internal standardization.}
SNR-preserving internal standardization rescales the entire raw observation
model:
\[
    \widetilde{\mathbf Y}^{\theta}(t)
    =
    D_\theta^{-1}\mathbf Y^\theta(t)
    =
    D_\theta^{-1}\mathbf X^\theta(t)
    +
    D_\theta^{-1}\boldsymbol\varepsilon(t).
\]
Hence
\[
    \widetilde{\mathbf Y}^{\theta}(t)
    =
    \widetilde{\mathbf X}^{\theta}(t)
    +
    \widetilde{\boldsymbol\varepsilon}^{\theta}(t),
\]
where
\[
    \widetilde{\boldsymbol\varepsilon}^{\theta}(t)
    =
    D_\theta^{-1}\boldsymbol\varepsilon(t).
\]
Because
\[
    \operatorname{Cov}\{\boldsymbol\varepsilon(t)\}
    =
    \sigma^2I_k,
\]
the standardized observation-noise covariance is
\[
    \operatorname{Cov}
    \{
        \widetilde{\boldsymbol\varepsilon}^{\theta}(t)
    \}
    =
    \sigma^2D_\theta^{-2}.
\]
Therefore the standardized observed covariance is
\begin{equation}
    \widetilde\Gamma_Y^\theta(\tau)
    =
    D_\theta^{-1}
    \Gamma_X^\theta(\tau)
    D_\theta^{-1}
    +
    \sigma^2D_\theta^{-2}
    \mathbf 1_{\{\tau=0\}} .
    \label{eq:snr_standardized_covariance}
\end{equation}
The effective standardized noise variance for task \(v\) is
\[
    \widetilde\sigma_v^2
    =
    \frac{\sigma^2}{q_v^\theta}.
\]
Consequently,
\[
    \operatorname{SNR}_v^{\rm std}
    =
    \frac{
        \operatorname{Var}\{\widetilde X_v^\theta(t)\}
    }{
        \operatorname{Var}\{\widetilde\varepsilon_v^\theta(t)\}
    }
    =
    \frac{1}{\sigma^2/q_v^\theta}
    =
    \frac{q_v^\theta}{\sigma^2}
    =
    \operatorname{SNR}_v^{\rm raw}.
\]
This justifies the term \emph{SNR-preserving internal standardization}.

We will use the following consequence of
Corollary~\ref{cor:raw_structgp_scale_invariance}: if two admissible raw
StructGP parameters \(\theta\) and \(\theta'\) induce latent covariance
functions that differ only by a global positive scale, that is, if
\[
    \Gamma_X^{\theta'}(\tau)
    =
    a^2\Gamma_X^\theta(\tau)
    \qquad
    \text{for all }\tau
\]
for some \(a>0\), then they induce the same StructGP ordered-dependence graph:
\[
    \mathcal G_H(\theta')
    =
    \mathcal G_H(\theta).
\]

\begin{assumption}[Separation of latent covariance and nugget]
\label{ass:nugget_separation_snr}
The observation design is rich enough to identify both the continuous latent
covariance term and the discontinuous nugget term in
\eqref{eq:snr_standardized_covariance}. Equivalently, equality of standardized
observation laws implies equality of the standardized latent covariance
functions and equality of the standardized nugget matrices.
\end{assumption}

\begin{theorem}[Ordered-graph identifiability under SNR-preserving internal standardization]
\label{thm:snr_preserving_identifiability}
Suppose that the raw StructGP model has shared task-independent observation
noise \(\sigma^2I_k\) with \(\sigma^2>0\), that
\(q_v^\theta>0\) for all tasks, and that
Assumption~\ref{ass:nugget_separation_snr} holds. Then the StructGP
ordered-dependence graph
\[
    \mathcal G_H(\theta)
    =
    \operatorname{supp}_{\rm off}H_\theta
    =
    \operatorname{supp}_{\rm off}S_\theta
\]
is identifiable from the SNR-preserving internally standardized observation
law of \(\widetilde{\mathbf Y}^{\theta}\).

More precisely, if two admissible parameters \(\theta\) and \(\theta'\) induce
the same SNR-preserving standardized observation law, then
\[
    \mathcal G_H(\theta)
    =
    \mathcal G_H(\theta').
\]
\end{theorem}

\begin{proof}
Let \(\theta\) and \(\theta'\) be two admissible raw StructGP
parameterizations, with latent covariance functions
\(\Gamma_X^\theta\) and \(\Gamma_X^{\theta'}\), and shared raw
observation-noise variances \(\sigma^2\) and \((\sigma')^2\),
respectively.
 Let
\[
    D_\theta
    =
    \operatorname{diag}(s_1,\ldots,s_k),
    \qquad
    D_{\theta'}
    =
    \operatorname{diag}(s'_1,\ldots,s'_k)
\]
be the corresponding internal scale matrices.

Assume that the two SNR-preserving standardized observation laws are equal.
Since the processes are Gaussian, equality of laws is equivalent to equality
of the standardized covariance functions:
\[
    \widetilde\Gamma_Y^\theta(\tau)
    =
    \widetilde\Gamma_Y^{\theta'}(\tau)
    \qquad
    \text{for all }\tau .
\]
By Assumption~\ref{ass:nugget_separation_snr}, equality of standardized
observation laws implies equality of both the standardized latent covariance
terms and the standardized nugget terms. Therefore,
\begin{equation}
    D_\theta^{-1}
    \Gamma_X^\theta(\tau)
    D_\theta^{-1}
    =
    D_{\theta'}^{-1}
    \Gamma_X^{\theta'}(\tau)
    D_{\theta'}^{-1}
    \qquad
    \text{for all }\tau .
    \label{eq:equal_standardized_latent_covariances_snr}
\end{equation}
and
\begin{equation}
    \sigma^2D_\theta^{-2}
    =
    (\sigma')^2D_{\theta'}^{-2}.
    \label{eq:equal_standardized_nuggets_snr}
\end{equation}

From \eqref{eq:equal_standardized_nuggets_snr}, for each task \(v\),
\[
    \frac{\sigma^2}{s_v^2}
    =
    \frac{(\sigma')^2}{(s'_v)^2}.
\]
Since \(\sigma,\sigma'>0\) and \(s_v,s'_v>0\), this gives
\[
    s'_v
    =
    a s_v
    \qquad
    \text{for all }v,
    \qquad
    a
    :=
    \frac{\sigma'}{\sigma}
    >
    0 .
\]
Hence
\[
    D_{\theta'}
    =
    aD_\theta .
\]

Substituting \(D_{\theta'}=aD_\theta\) into
\eqref{eq:equal_standardized_latent_covariances_snr} yields
\[
    D_\theta^{-1}
    \Gamma_X^\theta(\tau)
    D_\theta^{-1}
    =
    a^{-2}
    D_\theta^{-1}
    \Gamma_X^{\theta'}(\tau)
    D_\theta^{-1}.
\]
Multiplying on the left and right by \(D_\theta\), we obtain
\[
    \Gamma_X^{\theta'}(\tau)
    =
    a^2\Gamma_X^\theta(\tau)
    \qquad
    \text{for all }\tau .
\]
Thus the two raw latent covariance functions differ only by a global positive
scale.

By Corollary~\ref{cor:raw_structgp_scale_invariance}, global positive
rescaling of the raw latent covariance does not change the identified StructGP
ordered-dependence graph. Therefore,
\[
    \mathcal G_H(\theta')
    =
    \mathcal G_H(\theta).
\]
This proves identifiability of the StructGP ordered-dependence graph from the
SNR-preserving standardized observation law.
\end{proof}

\paragraph{Remark on the shared-noise assumption.}
The assumption that the raw observation-noise covariance is \(\sigma^2 I_k\) is
essential for the argument. Indeed, equality of standardized nuggets then
forces
\[
    \sigma^2D_\theta^{-2}
    =
    (\sigma')^2D_{\theta'}^{-2},
\]
and hence \(D_{\theta'}=aD_\theta\) for a single global scalar \(a>0\). If the
raw model allowed arbitrary task-specific noise variances, equality of
standardized nuggets would only identify taskwise ratios
\(\sigma_v^2/(s_v^\theta)^2\), and arbitrary taskwise rescalings would no longer
be ruled out without additional assumptions.

\subsection{Graph interpretation and identifiability in LP-StructGP}
\label{app:lp_structgp_graph_interpretation}

We now discuss how the ordered-dependence graph interpretation of
\textsc{StructGP} extends to \textsc{LP-StructGP}. The aim is not to prove
identifiability of all latent-pathway parameters. Indeed, the pathway labels,
subject-specific gates, and temporal shifts may be non-identifiable up to
mixture symmetries or other reparameterizations that leave the covariance
unchanged. Instead, we ask whether the shared inter-task graph encoded by
\(H\) retains its ordered conditional-independence interpretation and whether,
conditional on fixed subject-level pathway filters, it is identifiable from
the exact marginal likelihood.

\paragraph{Conditional stationarity with fixed gates and shifts.}
In \textsc{LP-StructGP}, each latent pathway is generated through the same
shared inter-task convolution filter:
\[
\mathbf Q_u(t)
=
(\mathbf H \ast \mathbf w_u)(t),
\qquad
u=1,\ldots,p,
\]
where \(\mathbf w_u\) is a \(k\)-dimensional white-noise process. A subject
trajectory is represented as a mixture of these pathways plus a subject-specific
residual component:
\begin{equation}
\mathbf Y_i(t)
=
\sum_{u=1}^p
(g_{iu} \ast \mathbf Q_u)(t)
+
(\mathbf H \ast \mathbf w_i)(t).
\label{eq:lp_structgp_subject_process}
\end{equation}
Here \(g_{iu}\) denotes the subject-specific pathway filter. In this subsection
we consider the separable case used in the LP-StructGP construction, where
\(g_{iu}\) is a scalar temporal filter acting identically on all task
components of the multivariate pathway \(\mathbf Q_u\). The subject-level
parameters include the pathway weights, length-scales, and temporal shifts.

For any fixed collection of subject-level filters, gates, and shifts, the
resulting LP-StructGP prior is stationary. Updating these parameters between
inference steps changes the parameter point at which the Gaussian process
likelihood is evaluated; it does not make the process associated with a fixed
parameter point time-varying.

The conditional identifiability result below uses a stronger form of this
conditioning: the same subject-level pathway filters are held fixed across the
competing shared inter-task parameterizations. It therefore does not establish
joint identifiability of the subject-level filters and the shared filter \(H\).

For example, if
\[
g_{iu}(t)
=
a_{iu}
\exp\!\left\{
-\frac{(t-\tau_{iu})^2}{\ell^{(\mathrm{sub})}_{iu}}
\right\},
\]
then its Fourier transform has the form
\[
\widetilde g_{iu}(\omega)
=
a_{iu}
e^{-i\omega\tau_{iu}}
\widetilde g^{\,0}_{iu}(\omega),
\]
where \(\widetilde g^{\,0}_{iu}\) is the Fourier transform of the corresponding
unshifted Gaussian filter. The shift therefore contributes only the phase factor
\(e^{-i\omega\tau_{iu}}\). It changes temporal alignment and cross-spectral
phase, but it does not introduce dependence on absolute time. Consequently,
conditional on fixed subject-level gates and shifts, the LP-StructGP prior is
stationary in time.

\paragraph{Preservation of the ordered conditional-independence graph.}
Taking the Fourier transform of equation~\eqref{eq:lp_structgp_subject_process}
gives
\[
\widetilde{\mathbf Y}_i(\omega)
=
\sum_{u=1}^p
\widetilde g_{iu}(\omega)
\widetilde{\mathbf H}(\omega)
\widetilde{\mathbf W}_u(\omega)
+
\widetilde{\mathbf H}(\omega)
\widetilde{\mathbf W}_i(\omega).
\]
Equivalently,
\[
\widetilde{\mathbf Y}_i(\omega)
=
\widetilde{\mathbf H}(\omega)
\left\{
\sum_{u=1}^p
\widetilde g_{iu}(\omega)
\widetilde{\mathbf W}_u(\omega)
+
\widetilde{\mathbf W}_i(\omega)
\right\}.
\]
Because the latent pathway noises and the subject-specific noises are mutually
independent, the within-subject task spectral density is
\begin{equation}
\mathbf F_i(\omega)
=
c_i(\omega)
\widetilde{\mathbf H}(\omega)
\widetilde{\mathbf H}(\omega)^* ,
\label{eq:lp_subject_spectral_density}
\end{equation}
where
\[
c_i(\omega)
=
1
+
\sum_{u=1}^p
\left\lvert
\widetilde g_{iu}(\omega)
\right\rvert^2 .
\]
The residual term implies that
\[
c_i(\omega)>0
\qquad
\text{for all } \omega .
\]
Thus, for each subject \(i\), the task spectral density is a positive
frequency-wise scalar multiple of the shared StructGP task spectral density.

A valid ordered lower-triangular spectral factor of \(\mathbf F_i(\omega)\) is
therefore
\[
\sqrt{c_i(\omega)}\,\widetilde{\mathbf H}(\omega).
\]
Since \(c_i(\omega)>0\), multiplication by \(\sqrt{c_i(\omega)}\) does not
change the zero pattern of the spectral factor:
\[
\widetilde H_{vu}(\omega)=0
\quad\Longleftrightarrow\quad
\sqrt{c_i(\omega)}\,
\widetilde H_{vu}(\omega)=0 .
\]
Consequently,
\[
H_{vu}\equiv 0
\quad\Longleftrightarrow\quad
\sqrt{c_i(\omega)}\,
\widetilde H_{vu}(\omega)=0
\quad
\text{for all } \omega .
\]
Hence the ordered conditional-independence interpretation of the sparsity
pattern of \(H\) is preserved in LP-StructGP, conditional on fixed
subject-level pathway parameters.
For ordered pairs \(u<v\), after fixing the task order in which
\(\widetilde{\mathbf H}(\omega)\) is lower triangular, this gives
\[
H_{vu}\equiv 0
\quad\Longleftrightarrow\quad
Y_{i,u}(\cdot)
\perp\!\!\!\perp
Y_{i,v}(\cdot)
\mid
\mathbf Y_{i,\{1,\ldots,u-1\}}(\cdot),
\]
where the conditional independence is understood in the ordered spectral
sense used for StructGP.

This is a preservation statement at a fixed parameter point: multiplication
by the specified positive function \(c_i(\omega)\) does not change the support
of the corresponding spectral factor. It does not by itself establish
identifiability across competing parameterizations in which
\(c_i(\omega)\) is also allowed to vary.

This preservation result relies on the separable pathway construction. If the
subject-pathway filters acted differently across tasks through non-scalar
task-specific filters, the subject-level layer could transform the task
spectral density by a non-scalar matrix. In that case, the ordered Cholesky
zeros of the observed subject-level spectrum would not necessarily coincide
with the zeros of \(H\). The graph would then remain a graph of the shared
latent inter-task mechanism, but not necessarily the ordered
conditional-independence graph of the marginal observed subject trajectory.

\paragraph{Conditional graph identifiability from the population marginal likelihood.}
Fix the collection of subject-level pathway filters
\(\{g_{iu}:i=1,\ldots,n,\ u=1,\ldots,p\}\), and compare only competing
parameterizations of the shared inter-task filter \(H\). Consequently, the
functions
\[
    c_i(\omega)
    =
    1+
    \sum_{u=1}^p
    \left|
        \widetilde g_{iu}(\omega)
    \right|^2
    >0
\]
are fixed across the candidate parameterizations.

Suppose that the observation design is sufficiently rich to identify the exact
LP-StructGP covariance law and that the observation-noise term is absent,
known, or separately identifiable. The population marginal law then identifies
the within-subject task spectral density \(\mathbf F_i(\omega)\). By
equation~\eqref{eq:lp_subject_spectral_density},
\[
    \mathbf F_i(\omega)
    =
    c_i(\omega)\mathbf F_H(\omega),
    \qquad
    \mathbf F_H(\omega)
    :=
    \widetilde{\mathbf H}(\omega)
    \widetilde{\mathbf H}(\omega)^* .
\]
Because \(c_i(\omega)\) is fixed and strictly positive, the shared StructGP
spectral density is identified as
\[
    \mathbf F_H(\omega)
    =
    \frac{1}{c_i(\omega)}
    \mathbf F_i(\omega).
\]
Proposition~\ref{prop:structgp_factor_uniqueness} can therefore be applied
directly to \(\mathbf F_H\). It identifies the anchored acyclic convolution
factor and, in particular, its off-diagonal support. Hence, conditional on the
fixed subject-level pathway filters, every population maximizer over the shared
StructGP parameters induces the same ordered-dependence graph
\[
    \mathcal G_H
    =
    \operatorname{supp}_{\rm off}H
    =
    \operatorname{supp}_{\rm off}S.
\]

This is a conditional graph-identifiability result. It does not assert joint
identifiability when the subject-level pathway filters are allowed to differ
between competing parameterizations. Indeed, if
\[
    c_i(\omega)\mathbf F_H(\omega)
    =
    c_i'(\omega)\mathbf F_H'(\omega),
\]
with both scalar functions unknown, the observation law determines only
\[
    \mathbf F_H'(\omega)
    =
    \frac{c_i(\omega)}{c_i'(\omega)}
    \mathbf F_H(\omega).
\]
This frequency-dependent rescaling is not the global constant rescaling
covered by Corollary~\ref{cor:raw_structgp_scale_invariance}; it may alter the
diagonal spectral shapes used in identifying the anchored factor. No
unconditional graph-identifiability claim is therefore made when the
subject-level filters and their length-scales are jointly varied.

\paragraph{Cohort-conditioned predictions and nonstationarity.}
The stationarity argument above concerns the generative LP-StructGP prior
conditional on fixed subject-level parameters. After observing a training
cohort, the predictive law used for a new subject is different. The latent
pathways are no longer represented only by their zero-mean stationary prior;
instead, their distribution is conditioned on the training data.

Let \(\mathcal D_{\mathrm{train}}\) denote the observed training cohort. Even if
the prior pathway covariance is stationary, the posterior pathway process
\[
\mathbf Q_u
\mid
\mathcal D_{\mathrm{train}}
\]
generally has a nonzero mean
\[
\mathbf m_u(t)
=
\mathbb E
\left[
\mathbf Q_u(t)
\mid
\mathcal D_{\mathrm{train}}
\right],
\]
and a posterior covariance
\[
\mathbf K^{\mathrm{post}}_{uu'}(t,t')
=
\operatorname{Cov}
\left\{
\mathbf Q_u(t),
\mathbf Q_{u'}(t')
\mid
\mathcal D_{\mathrm{train}}
\right\}.
\]
For Gaussian processes, posterior covariances after conditioning on finite
observations generally depend on the absolute time points \(t\) and \(t'\), not
only on their lag \(t-t'\). Hence the cohort-conditioned pathway posterior is,
in general, nonstationary.

For a new subject with fixed current pathway parameters, the cohort-conditioned
predictive mean is
\[
\mathbf m_i(t)
=
\sum_{u=1}^p
(g_{iu} \ast \mathbf m_u)(t).
\]
The predictive covariance contains both the subject-specific residual covariance
and the pathway-posterior covariance:
\[
\begin{aligned}
\mathbf K_i^{\mathrm{pred}}(t,t')
&=
\mathbf K_{\mathrm{res}}(t-t')
\\
&\quad+
\sum_{u=1}^p
\sum_{u'=1}^p
\int_{\mathbb R}
\int_{\mathbb R}
g_{iu}(t-s)
\,\mathbf K^{\mathrm{post}}_{uu'}(s,s')\,
g_{iu'}(t'-s')
\, ds\, ds' .
\end{aligned}
\]
The residual term \(\mathbf K_{\mathrm{res}}(t-t')\) is stationary, but the
pathway-posterior term need not be. As a result, the predictive mean and
predictive covariance for a new subject may exhibit temporal trends, recovery
patterns, deterioration patterns, and time-varying uncertainty.

Thus LP-StructGP combines a stationary convolutional generative mechanism with
a cohort-conditioned, generally nonstationary posterior predictive law. The
graph interpretation should therefore be attached to the shared stationary
mechanism \(H\), not to the full finite-sample posterior predictive covariance
after conditioning on the training cohort. The learned graph describes the
shared ordered inter-task dependency structure used to generate both the latent
pathways and the subject-specific residual component. The latent-pathway layer
then uses the training cohort to construct data-conditioned trajectory
templates, which can yield nonzero and nonstationary patient-specific
predictive means and covariances without introducing a separate
pathway-specific task graph.

\subsection{Stability of the detached accumulator and implications for
optimization stationarity and latent-pathway recovery}
\label{app:detached_accumulator_stability}

This subsection analyzes the stability of the detached, geometrically faded
accumulator used for scalable \textsc{LP-StructGP} training. We first compare
the implemented accumulator with a fixed-current reference state in which the
retained sufficient statistics are recomputed at the current parameter value,
and show how their discrepancy is controlled by parameter variation over the
effective memory window. We then discuss the additional optimization and
statistical assumptions that would be required to extend this stability
analysis to approximate optimization stationarity and latent-pathway recovery.

\paragraph{Finite-rank latent-pathway representation.}

After the HSGP approximation, the shared latent-pathway component can be written
as a finite-dimensional Gaussian latent-variable model:
\[
    a\sim\mathcal N(0,I_q),
    \qquad
    y_i\mid a,\theta
    \sim
    \mathcal N\!\left(
        \Phi_i(\theta)a,
        M_i(\theta)
    \right),
\]
where \(a\in\mathbb R^q\) contains the retained HSGP coefficients of the shared
latent pathways. The feature matrix \(\Phi_i(\theta)\) may depend on the
inter-task convolution parameters, the subject--pathway coupling filters,
temporal shifts, and softmax gating weights. The gates are treated as
differentiable components of \(\theta\), rather than as latent categorical
assignments.

For a fixed parameter value \(\theta\), define the local Woodbury statistics
\[
    s_i(\theta)
    =
    y_i^\top M_i(\theta)^{-1}y_i,
\]
\[
    d_i(\theta)
    =
    \Phi_i(\theta)^\top M_i(\theta)^{-1}y_i,
\]
and
\[
    c_i(\theta)
    =
    \Phi_i(\theta)^\top
    M_i(\theta)^{-1}
    \Phi_i(\theta).
\]
For mini-batch \(\mathcal B_t\), let
\[
    z_t(\theta)
    =
    \big(
        s_t(\theta),
        d_t(\theta),
        c_t(\theta)
    \big),
\]
where
\[
    s_t(\theta)
    =
    \sum_{i\in\mathcal B_t}s_i(\theta),
    \qquad
    d_t(\theta)
    =
    \sum_{i\in\mathcal B_t}d_i(\theta),
\]
and
\[
    c_t(\theta)
    =
    \sum_{i\in\mathcal B_t}c_i(\theta).
\]
The corresponding accumulator state is denoted
\[
    Z_t=(S_t,D_t,C_t).
\]

For a fixed parameter value and a fixed collection of observations, the
posterior distribution of \(a\) is Gaussian:
\[
    a\mid y,\theta
    \sim
    \mathcal N(\mu_\theta,\Sigma_\theta),
\]
with
\[
    \Sigma_\theta
    =
    (I+C_\theta)^{-1},
    \qquad
    \mu_\theta
    =
    \Sigma_\theta D_\theta.
\]
Thus \(C_\theta\) contributes to the posterior precision, while \(D_\theta\)
is the corresponding natural parameter.

\paragraph{Exact finite-rank HSGP reference target.}
Let
\[
    \mathcal L_{\mathrm{HSGP}}(\theta)
\]
denote the exact negative marginal likelihood of the finite-rank HSGP model,
obtained by analytically integrating out the latent coefficients \(a\). As
shown in Section~\ref{app:minibatch_nmll}, when \(\theta\) is fixed during a
complete sweep and no geometric fading is applied, the conditional Woodbury
increments telescope:
\[
    \sum_{t=1}^{T}
    \mathcal L_t(\theta)
    =
    \mathcal L_{\mathrm{HSGP}}(\theta).
\]
The resulting objective is therefore exact for the finite-rank model; its only
model-level approximation relative to the original GP is the HSGP truncation.

We use \(\theta^\star\) to denote a reference population parameter, or a
representative of the population minimizer equivalence class, for this
finite-rank model. This reference is used only to discuss the behavior of the
detached accumulator after the parameter sequence has stabilized.

\paragraph{Detached faded state and fixed-current reference state.}

In the implementation, the accumulator is updated using geometric fading:
\[
    \bar Z_t
    =
    \beta\,\operatorname{sg}(\bar Z_{t-1})
    +
    z_t(\theta_t),
    \qquad
    0\leq\beta<1,
\]
where \(\operatorname{sg}\) denotes the stop-gradient operator. Only the
pre-update historical state is detached; gradients are retained through the
current-batch statistics \(z_t(\theta_t)\).

Componentwise,
\[
    \bar S_t
    =
    \beta\,\operatorname{sg}(\bar S_{t-1})
    +
    s_t(\theta_t),
\]
\[
    \bar D_t
    =
    \beta\,\operatorname{sg}(\bar D_{t-1})
    +
    d_t(\theta_t),
\]
and
\[
    \bar C_t
    =
    \beta\,\operatorname{sg}(\bar C_{t-1})
    +
    c_t(\theta_t).
\]
The associated faded Gaussian state is
\[
    \bar\Sigma_t
    =
    (I+\bar C_t)^{-1},
    \qquad
    \bar\mu_t
    =
    \bar\Sigma_t\bar D_t.
\]
For fixed \(\beta<1\), this is a discounted Gaussian state rather than the
exact full-data posterior state.

Assuming zero initialization and batches indexed from \(1\), the numerical
state can be written as
\[
    \bar Z_t
    =
    \sum_{\ell=0}^{t-1}
    \beta^\ell
    z_{t-\ell}(\theta_{t-\ell}).
\]
To separate fading from parameter staleness, define the fixed-current state
\[
    Z_t^\beta(\theta_t)
    =
    \sum_{\ell=0}^{t-1}
    \beta^\ell
    z_{t-\ell}(\theta_t),
\]
which uses the same geometrically weighted batches, but recomputes every
retained statistic at the current parameter value.

Equivalently, define the fixed-current historical state
\[
    Z_{t-1}^{\beta,\mathrm{hist}}(\theta_t)
    =
    \sum_{\ell=0}^{t-2}
    \beta^\ell
    z_{t-1-\ell}(\theta_t),
\]
with the empty sum equal to zero for \(t=1\). Then
\[
    Z_t^\beta(\theta_t)
    =
    \beta Z_{t-1}^{\beta,\mathrm{hist}}(\theta_t)
    +
    z_t(\theta_t).
\]
The comparison between \(\bar Z_t\) and \(Z_t^\beta(\theta_t)\) therefore
isolates the effect of retaining historical statistics evaluated at earlier
parameter values. It does not include the separate approximation caused by
finite geometric memory.

\paragraph{Control of the effective-memory parameter drift.}

Define the effective-memory parameter drift
\[
    \delta_t^{(\beta)}
    =
    (1-\beta)
    \sum_{\ell=0}^{t-1}
    \beta^\ell
    \|\theta_t-\theta_{t-\ell}\|,
\]
where \(\|\cdot\|\) denotes any fixed norm on the finite-dimensional parameter
or accumulator space. This quantity measures how far the current parameter has
moved from the parameter values used to construct the retained statistics.

Suppose locally that the mini-batch statistics are Lipschitz in the parameter:
\[
    \|z_s(\theta)-z_s(\theta')\|
    \leq
    L_z\|\theta-\theta'\|.
\]
Using the definitions of the implemented state and the fixed-current state,
\[
    \bar Z_t
    =
    \sum_{\ell=0}^{t-1}
    \beta^\ell
    z_{t-\ell}(\theta_{t-\ell}),
    \qquad
    Z_t^\beta(\theta_t)
    =
    \sum_{\ell=0}^{t-1}
    \beta^\ell
    z_{t-\ell}(\theta_t),
\]
their difference satisfies
\[
\begin{aligned}
    \left\|
        \bar Z_t-Z_t^\beta(\theta_t)
    \right\|
    &=
    \left\|
        \sum_{\ell=0}^{t-1}
        \beta^\ell
        \left[
            z_{t-\ell}(\theta_{t-\ell})
            -
            z_{t-\ell}(\theta_t)
        \right]
    \right\|
    &&\text{by the definitions of \(\bar Z_t\) and \(Z_t^\beta(\theta_t)\)}
    \\
    &\leq
    \sum_{\ell=0}^{t-1}
    \beta^\ell
    \left\|
        z_{t-\ell}(\theta_{t-\ell})
        -
        z_{t-\ell}(\theta_t)
    \right\|
    &&\text{by the triangle inequality and \(\beta^\ell\geq0\)}
    \\
    &\leq
    L_z
    \sum_{\ell=0}^{t-1}
    \beta^\ell
    \|\theta_{t-\ell}-\theta_t\|
    &&\text{by the local Lipschitz continuity of \(z_{t-\ell}\)}
    \\
    &=
    \frac{L_z}{1-\beta}
    \delta_t^{(\beta)}
    &&\text{by the definition of \(\delta_t^{(\beta)}\).}
\end{aligned}
\]

Thus the stale-state discrepancy is controlled by the parameter motion over
the effective memory window. For fixed \(\beta<1\), if the parameter sequence
varies little over that window, then the implemented accumulator is close to
the state that would be obtained by recomputing all retained statistics at the
current parameter value. If the conditional loss is also locally Lipschitz in
the accumulator state, with constant \(L_{\mathcal F}\), then
\[
    \left|
        \mathcal F_t(\theta_t,\bar Z_t)
        -
        \mathcal F_t\!\left(
            \theta_t,Z_t^\beta(\theta_t)
        \right)
    \right|
    \leq
    \frac{L_{\mathcal F}L_z}{1-\beta}
    \delta_t^{(\beta)}.
\]

The evolution of the effective-memory drift can be related directly to the
parameter update. For
\[
    \theta_{t+1}
    =
    \theta_t-\eta_t\widehat g_t,
\]
we have
\[
\begin{aligned}
    \delta_{t+1}^{(\beta)}
    &=
    (1-\beta)
    \sum_{\ell=0}^{t}
    \beta^\ell
    \|\theta_{t+1}-\theta_{t+1-\ell}\|
    \\
    &=
    \beta(1-\beta)
    \sum_{m=0}^{t-1}
    \beta^m
    \|\theta_{t+1}-\theta_{t-m}\|,
\end{aligned}
\]
because the term corresponding to \(\ell=0\) is zero. By the triangle
inequality,
\[
    \|\theta_{t+1}-\theta_{t-m}\|
    \leq
    \|\theta_t-\theta_{t-m}\|
    +
    \|\theta_{t+1}-\theta_t\|.
\]
Therefore,
\[
\begin{aligned}
    \delta_{t+1}^{(\beta)}
    &\leq
    \beta\delta_t^{(\beta)}
    +
    \beta(1-\beta^t)
    \|\theta_{t+1}-\theta_t\|
    \\
    &\leq
    \beta\delta_t^{(\beta)}
    +
    \beta\eta_t\|\widehat g_t\|.
\end{aligned}
\]

This recursion separates two effects. The term
\(\beta\delta_t^{(\beta)}\) represents the contraction produced by geometric
forgetting: previous parameter differences become one step older and receive
an additional factor \(\beta\). The term
\(\beta\eta_t\|\widehat g_t\|\) is the new drift introduced by the latest
parameter update. In particular, if the parameter does not move, then
\[
    \delta_{t+1}^{(\beta)}
    \leq
    \beta\delta_t^{(\beta)},
\]
so the stale-parameter drift contracts geometrically. More generally, the
drift decreases whenever the new parameter step is sufficiently small
relative to the existing drift, for example when
\[
    \eta_t\|\widehat g_t\|
    <
    \frac{1-\beta}{\beta}
    \delta_t^{(\beta)}.
\]

If the stochastic gradients are uniformly bounded,
\(\|\widehat g_t\|\leq G\), and the learning rate is constant
\(\eta_t=\eta\), the recursion gives the residual bound
\[
    \limsup_{t\to\infty}
    \delta_t^{(\beta)}
    \leq
    \frac{\beta\eta G}{1-\beta}
    =
    O\!\left(
        \frac{\eta}{1-\beta}
    \right).
\]
The corresponding accumulator-state discrepancy is then bounded at the
larger order
\[
    O\!\left(
        \frac{\eta}{(1-\beta)^2}
    \right)
\]
through the preceding Lipschitz bound. By contrast, if
\(\eta_t\|\widehat g_t\|\to0\), the forcing term in the drift recursion
vanishes, and the stale-parameter component may converge to zero. These
relations describe the stability of the detached accumulator relative to its
fixed-current counterpart; they do not, by themselves, establish convergence
of the parameter sequence or stationarity of the optimization objective.

\paragraph{Discussion of optimization stationarity and latent-pathway recovery.}

The preceding bounds establish control of the discrepancy between the
implemented detached accumulator and the faded state that would be obtained by
recomputing the retained statistics at the current parameter value. They do
not, by themselves, establish convergence of the optimization procedure or
statistical recovery of the latent pathways. These stronger conclusions would
require additional assumptions.

First, an approximate-stationarity result would require standard
stochastic-approximation conditions, including bounded iterates, a suitable
learning-rate schedule, bounded gradient variance, and asymptotically
negligible optimization bias. The stale-state discrepancy analyzed above is
one source of such bias. A complete analysis would also need to account for
the fact that the retained historical state is treated as detached when the
parameter update is computed, and to characterize the limiting expected
detached update. If this expected update corresponds to the gradient of a
faded surrogate objective \(\Psi_\beta\), and if
\[
    \sup_{\theta\in\Theta}
    \left\|
        \nabla\Psi_\beta(\theta)
        -
        \nabla\mathcal L_{\mathrm{HSGP}}(\theta)
    \right\|
    \leq
    \varepsilon_\beta^{\mathrm{sur}},
\]
then any stationary point \(\bar\theta\) of the surrogate would satisfy
\[
    \left\|
        \nabla\mathcal L_{\mathrm{HSGP}}(\bar\theta)
    \right\|
    \leq
    \varepsilon_\beta^{\mathrm{sur}}.
\]
Thus convergence to the surrogate stationary set would imply approximate
stationarity for the finite-rank HSGP objective only when the combined
discrepancy introduced by fading, stale-state evaluation, and detachment is
sufficiently small.

Second, accumulator stability does not by itself imply recovery of the latent
pathways. Such a result would additionally require the parameter sequence to
approach the statistically correct equivalence class, up to pathway-label
permutations or other covariance-preserving reparameterizations, and the
retained observations to provide sufficient information in every identifiable
latent direction. To illustrate the relevant statistical mechanism, consider
distinct subject-level blocks that are conditionally independent given latent
coefficients and define
\[
    C_{\mathrm{eff}}
    =
    \sum_i
    w_i
    \Phi_i(\theta^\star)^\top
    M_i(\theta^\star)^{-1}
    \Phi_i(\theta^\star),
\]
and
\[
    D_{\mathrm{eff}}
    =
    \sum_i
    w_i
    \Phi_i(\theta^\star)^\top
    M_i(\theta^\star)^{-1}y_i.
\]
Under the model
\[
    y_i
    =
    \Phi_i(\theta^\star)a^\star+\varepsilon_i,
    \qquad
    \varepsilon_i
    \overset{\mathrm{ind}}{\sim}
    \mathcal N\!\left(
        0,M_i(\theta^\star)
    \right),
\]
the corresponding Gaussian-state mean satisfies
\[
\begin{aligned}
    (I+C_{\mathrm{eff}})^{-1}D_{\mathrm{eff}}-a^\star
    &=
    -(I+C_{\mathrm{eff}})^{-1}a^\star
    \\
    &\quad+
    (I+C_{\mathrm{eff}})^{-1}
    \sum_i
    w_i
    \Phi_i(\theta^\star)^\top
    M_i(\theta^\star)^{-1}
    \varepsilon_i.
\end{aligned}
\]
The first term represents Gaussian-prior shrinkage, whereas the second
represents weighted statistical error. This decomposition suggests that
latent-state concentration requires both increasing information in all
relevant latent directions and sufficient control of the weighted noise
contribution. If the effective information remains bounded, only finite
statistical resolution should generally be expected.

For fixed geometric fading over a stream of distinct, comparably informative
subject blocks, the effective information scale is of order
\[
    \sum_i w_i
    \asymp
    \frac{1}{1-\beta}.
\]
Fixed \(\beta<1\) therefore imposes a finite effective memory and need not yield
exact concentration. Stronger recovery guarantees would require increasing
information from distinct observations, for example in a no-fading regime or
under a separately analyzed growing-memory schedule, together with adequate
control of the stale-state and optimization biases. Repeated optimization
passes over the same finite cohort do not constitute additional independent
statistical information.

Accordingly, contraction of the effective parameter drift should be interpreted
as a stability property of the detached accumulator. Extending this result to
full optimization convergence or latent-pathway recovery would require a
separate analysis of the stochastic optimization dynamics, the surrogate-update
bias, parameter identifiability, weighted statistical error, and information
growth.

\clearpage
\bibliography{structgp}